\documentclass[letterpaper]{article} 
\usepackage{aaai25}  
\usepackage{times}  
\usepackage{helvet}  
\usepackage{courier}  
\usepackage[hyphens]{url}  
\usepackage{graphicx} 
\usepackage{natbib}  
\usepackage{amsfonts}
\usepackage{caption} 
\usepackage{algorithm}
\usepackage{algorithmic}
\usepackage[toc,page]{appendix}

\usepackage{placeins}

\usepackage{newfloat}
\usepackage{listings}
\DeclareCaptionStyle{ruled}{labelfont=normalfont,labelsep=colon,strut=off} 
\floatstyle{ruled}
\newfloat{listing}{tb}{lst}{}
\floatname{listing}{Listing}
\title{CiTrus: Squeezing Extra Performance out of Low-data Bio-signal Transfer Learning}
\author {
    Eloy Geenjaar\textsuperscript{\rm 1}\thanks{Work completed during an internship at Dolby Laboratories},      
    Lie Lu\textsuperscript{\rm 2}
}
\affiliations {
    \textsuperscript{\rm 1}Georgia Institute of Technology,   
    \textsuperscript{\rm 2}Dolby Laboratories\\
    egeenjaar@gatech.edu, llu@dolby.com
}

\def\doubleunderline#1{\underline{\underline{#1}}}

\usepackage{bibentry}

\begin{document}

\maketitle

\begin{abstract}
 Transfer learning for bio-signals has recently become an important technique to improve prediction performance on downstream tasks with small bio-signal datasets. Recent works have shown that pre-training a neural network model on a large dataset (e.g. EEG) with a self-supervised task, replacing the self-supervised head with a linear classification head, and fine-tuning the model on different downstream bio-signal datasets (e.g., EMG or ECG) can dramatically improve the performance on those datasets. In this paper, we propose a new convolution-transformer hybrid model architecture with masked auto-encoding for low-data bio-signal transfer learning, introduce a frequency-based masked auto-encoding task, employ a more comprehensive evaluation framework, and evaluate how much and when (multimodal) pre-training improves fine-tuning performance. We also introduce a dramatically more performant method of aligning a downstream dataset with a different temporal length and sampling rate to the original pre-training dataset. Our findings indicate that the convolution-only part of our hybrid model can achieve state-of-the-art performance on some low-data downstream tasks. The performance is often improved even further with our full model. In the case of transformer-based models we find that pre-training especially improves performance on downstream datasets, multimodal pre-training often increases those gains further, and our frequency-based pre-training performs the best on average for the lowest and highest data regimes.
\end{abstract}

\begin{figure*}[t]
\centering
\includegraphics[width=\textwidth]{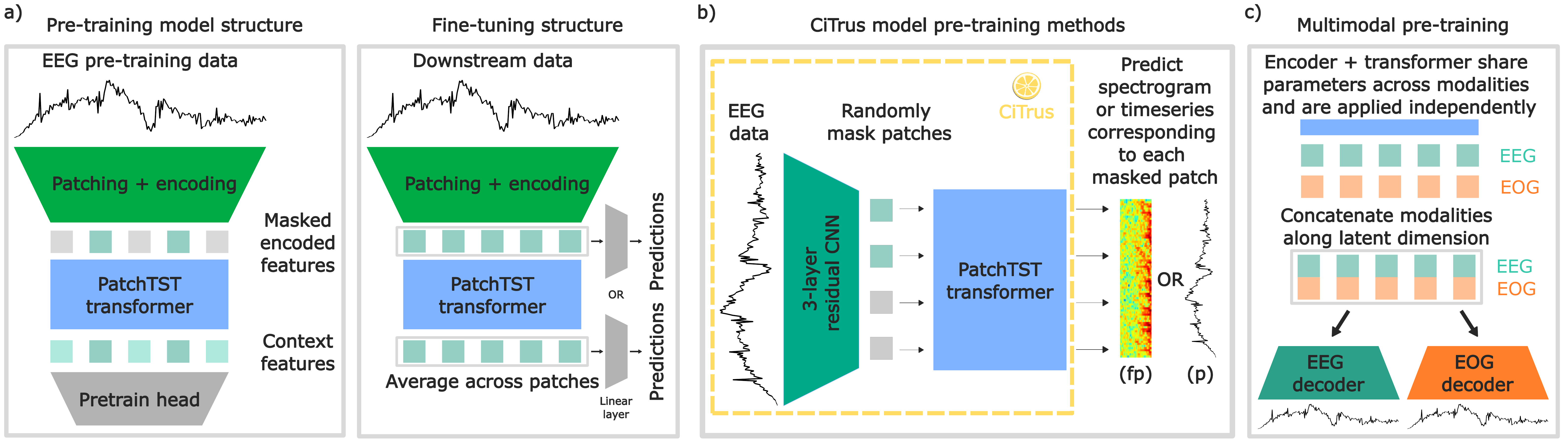}
\caption{Subfigure a) shows the general transfer learning framework; the pre-training and fine-tuning structures. Subfigure b) shows the pre-training structure of our proposed model, and sub-figure c) shows how the structure of the model is adapted to accommodate multi-modal pre-training data.}
\label{fig:model}
\end{figure*}

\section{Introduction}
The wearable market is growing quickly around the world~\cite{casselman2017wearable}. This increase in wearable usage means there is an increasing amount of bio-signal data available that can be used in the field of preventative medicine. By combining subjective assessments of experiences with objective signals extracted from bio-signals related to someone’s well-being, stress level, cognitive state, etc., doctors can make more informed treatment decisions. There are many types of bio-signals, each recording a different type of information, for example, electroencephalography (EEG) non-invasively records the brain’s electrical activity from outside the skull and can be used to predict attention levels~\cite{li2011real}. On the other hand, photoplethysmography (PPG) records volumetric blood changes, and can be used to predict stress levels~\cite{charlton2018assessing} from non-invasive recordings on the skin. Many more bio-signals exist, each with unique information about a potential patient.

However, bio-signals also come with three important drawbacks. First, bio-signals are noisy; with EEG, for example, the electrical activity from the brain is recorded through the skull and must thus travel through a dense bone before it arrives at the electrodes, which leads to noise. Moreover, it is harder to record activity deeper in the brain because the further away electrical current is generated from the electrode, the harder it is to reliably capture it. Additionally, normal activities while wearing wearables, such as walking, can induce movement-related noise in non-invasive bio-signals. Secondly, bio-signals suffer from a lot of subject-variability, i.e., it can be hard to generalize predictions from one person’s bio-signal to the same bio-signal recorded from another person. Lastly, making predictions about a person’s internal state from bio-signals is often non-trivial. These predictions require complex and non-linear transformations of the original bio-signal. This has led to the wide-scale adoption of neural networks in bio-signal research. However, neural networks require substantial amounts of labeled data to train on, and labeling bio-signals is expensive because it requires (medical) experts to go through the data and label the individual time windows of each bio-signal. It is thus imperative to develop neural network models that do not require many labeled examples to train on.

Transfer learning mitigates the need for many examples by first pre-training a neural network on a large unlabeled dataset. Transfer learning has shown to be effective for bio-signals with models pre-trained using a self-supervised task~\cite{zhang2022self,liu2023frequency,dong2024simmtm}. In this work, we develop a comprehensive evaluation strategy for bio-signal transfer learning and find that convolution-based architectures often outperform transformer-only models. We thus propose a new convolution transformer hybrid model that we call CiTrus and new pre-training and fine-tuning strategies. Our model systematically improves performance over previous methods and strong baselines we introduce for comparison in this paper.

In summary, our main contributions are:
\begin{itemize}
    \item The introduction of a convolution and transformer hybrid for bio-signal transfer learning (CiTrus) that outperforms previous models by a significant margin.
    \item Two pre-training approaches that both significantly improve fine-tuning performance for our propsed model.
    \item A new transferring approach that keeps the sampling frequency the same between the pre-training and fine-tuning dataset, which for most datasets, significantly improves performance.
    \item A more comprehensive evaluation of transfer learning models for bio-signals. We add new datasets, evaluate models over multiple data availability regimes, and use multiple test splits. We find that each of these is important for a fair comparison of bio-signal transfer learning.
\end{itemize}

\begin{table*}[!t]
    \begin{tabular}{l|ccc|ccc|ccc|ccc|}
    \hline 
 Dataset & \multicolumn{3}{c}{EMG@20\%} & \multicolumn{3}{c}{ECG@0.5\%} & \multicolumn{3}{c}{PPG@10\%} & \multicolumn{3}{c}{HAR@10\%} \\ 
 \hline 
 Metric & ACC & ROC & PRC & ACC & ROC & PRC & ACC & ROC & PRC & ACC & ROC & PRC \\ 
 \hline 
PatchTST (s) & 51.46 & 61.87 & 52.07 & 54.15 & 52.34 & 60.85 & 59.98 & 73.89 & 55.32 & 67.67 & 88.62 & 69.83 \\ 
PatchTST (p) & 57.93 & 71.35 & 62.46 & 56.63 & 52.89 & 61.34 & 59.74 & 74.57 & 56.31 & 60.57 & 85.85 & 62.7 \\ 
bioFAME (s) & 50.43 & 61.89 & 53.46 & 58.36 & 51.66 & 60.52 & 60.31 & 73.98 & 55.2 & 54.03 & 84.31 & 56.72 \\ 
bioFAME (mp) & 70.93 & 86.94 & 78.78 & 64.26 & 52.89 & 61.36 & 60.21 & \underline{76.67} & 58.51 & 60.5 & 86.85 & 65.54 \\ 
NLPatchTST (s) & 64.64 & 81.15 & 71.84 & 52.28 & 51.92 & 60.53 & 56.81 & 72.27 & 53.42 & 68.54 & 88.94 & 70.01 \\ 
NLPatchTST (p) & 70.45 & 87.9 & 80.11 & 54.43 & 52.96 & 61.27 & 60.39 & 75.15 & 56.62 & 65.66 & 87.78 & 67.61 \\ 
NLPatchTST (mp) & 74.61 & 90.16 & 82.77 & 56.86 & 53.02 & 61.41 & 59.44 & 74.78 & 56.14 & 60.41 & 86.2 & 63.7 \\ 
SimMTM (s) & 52.73 & 84.79 & 75.52 & 58.47 & 53.56 & 61.86 & 54.53 & \doubleunderline{76.83} & \doubleunderline{59.49} & \underline{75.41} & \underline{91.14} & \underline{78.35} \\ 
SimMTM (p) & 42.52 & 61.3 & 54.13 & 56.77 & 53.08 & 61.5 & 50.58 & 73.7 & 56.07 & \doubleunderline{77.37} & \doubleunderline{91.76} & \doubleunderline{80.14} \\ 
Ci (s) & \doubleunderline{96.55} & \doubleunderline{99.1} & \underline{97.99} & 66.89 & \underline{55.33} & \underline{63.64} & 58.91 & 72.12 & 54.47 & \textbf{80.71} & \textbf{93.03} & \textbf{82.8} \\ 
Ci (p) & 94.74 & 98.72 & 96.73 & 65.62 & 54.76 & 63.07 & 58.57 & 73.85 & 55.77 & 70.7 & 90.18 & 74.46 \\ 
CiTrus (s) & \underline{95.84} & \textbf{99.48} & \textbf{99.4} & \doubleunderline{69.35} & \doubleunderline{55.51} & \doubleunderline{63.78} & 59.18 & 72.35 & 56.03 & 71.87 & 89.97 & 74.89 \\ 
CiTrus (p) & 90.45 & 96.96 & 94.18 & 64.55 & 54.6 & 62.84 & \underline{62.41} & 75.37 & \underline{59.14} & 73.89 & 90.9 & 75.9 \\ 
CiTrus (fp) & \textbf{97.92} & \underline{99.02} & \doubleunderline{98.76} & \textbf{82.44} & \textbf{55.9} & \textbf{64.3} & \textbf{65.25} & \textbf{79.45} & \textbf{63.65} & 65.3 & 87.55 & 68.75 \\ 
CiTrus (mp) & 92.9 & 99.01 & 96.86 & \underline{67.72} & 55.2 & 63.6 & \doubleunderline{62.98} & 76.15 & 58.64 & 72.78 & 90.47 & 75.32 \\ 
\hline 
 Dataset & \multicolumn{3}{c}{FDB@0.5\%} & \multicolumn{3}{c}{Epilepsy@1\%} & \multicolumn{3}{c}{Gesture@10\%} & \multicolumn{3}{c}{SleepEDF@0.5\%} \\ 
 \hline 
 Metric & ACC & ROC & PRC & ACC & ROC & PRC & ACC & ROC & PRC & ACC & ROC & PRC \\ 
 \hline 
PatchTST (s) & 52.99 & 54.35 & 38.04 & 92.79 & 92.8 & 96.54 & 43.21 & 81.04 & 49.29 & 64.06 & 65.54 & 33.83 \\ 
PatchTST (p) & 60.15 & 55.91 & 39.59 & 93.34 & 93.81 & 97.09 & 49.6 & 83.79 & 56.36 & \underline{71.05} & 67.8 & 36.33 \\ 
bioFAME (s) & 52.07 & 53.63 & 37.13 & 92.46 & 96.17 & 98.72 & 30.36 & 74.46 & 39.54 & 70.62 & 68.04 & 37.45 \\ 
bioFAME (mp) & 56.71 & 56.64 & 39.74 & 92.77 & 96.93 & 99.01 & 36.38 & 79.27 & 46.28 & 70.51 & 68.1 & 37.91 \\ 
NLPatchTST (s) & 58.61 & 56.81 & 39.68 & 84.03 & 78.38 & 88.43 & 43.35 & 82.06 & 50.31 & 58.59 & 64.53 & 33.79 \\ 
NLPatchTST (p) & 63.64 & 57.28 & 40.24 & 92.64 & 92.41 & 96.15 & 48.13 & \underline{84.6} & 56.8 & \textbf{74.08} & 68.16 & 36.74 \\ 
NLPatchTST (mp) & 65.01 & 58.2 & 41.03 & 92.77 & 92.33 & 96.23 & \underline{51.88} & 84.2 & \underline{57.39} & 66.98 & 66.8 & 34.94 \\ 
SimMTM (s) & 43.26 & 52.69 & 36.45 & 93.35 & 55.15 & 86.21 & \doubleunderline{53.26} & \doubleunderline{84.7} & \textbf{63.2} &  &  &  \\ 
SimMTM (p) & 68.89 & 60.59 & 44.66 & 93.21 & 55.03 & 85.63 & \textbf{56.83} & \textbf{85.43} & \doubleunderline{63.0} &  &  &  \\ 
Ci (s) & \underline{74.26} & \doubleunderline{65.26} & \doubleunderline{50.37} & 93.75 & \doubleunderline{97.74} & \doubleunderline{99.31} & 35.09 & 75.98 & 43.72 & 69.02 & \textbf{69.03} & \textbf{38.84} \\ 
Ci (p) & 72.72 & \underline{64.45} & \underline{48.88} & \doubleunderline{94.37} & \textbf{97.93} & \textbf{99.36} & 40.31 & 81.15 & 49.04 & 70.48 & \underline{68.58} & \doubleunderline{38.57} \\ 
CiTrus (s) & 72.34 & 63.87 & 47.85 & 93.2 & \underline{97.68} & \underline{99.17} & 42.77 & 81.6 & 51.08 & 67.61 & 67.75 & 36.41 \\ 
CiTrus (p) & \doubleunderline{75.48} & 63.52 & 47.69 & \underline{94.08} & 95.29 & 97.92 & 40.71 & 80.73 & 50.63 & \doubleunderline{73.87} & \doubleunderline{68.85} & \underline{37.96} \\ 
CiTrus (fp) & \textbf{79.78} & \textbf{66.63} & \textbf{52.35} & 93.41 & 95.83 & 98.26 & 51.34 & 82.28 & 55.81 & 48.69 & 58.11 & 27.52 \\ 
CiTrus (mp) & 73.55 & 63.63 & 47.77 & \textbf{94.53} & 94.67 & 97.34 & 47.99 & 83.24 & 55.99 & 69.26 & 67.16 & 35.84 \\ 
\hline
\end{tabular}
\caption{A comparison of the different model architectures. We use the same evaluation method for each model, for EMG and FD-B we interpolate the data for fine-tuning, and for the other datasets we use our sliding window approach. The letters in brackets refer to how the model is trained; (s) is trained from scratch, (p) means it is pre-trained, (mp) means it uses multi-modal pre-training, and (fp) means it uses frequency pre-training. ACC, ROC, and PRC refer to the accuracy, area under the receiver operating characteristic, and the area under the precision recall curve, respectively. The best result for each metric is shown in bold, the second best result is double-underlined, and the third best result is single-underlined.}
\label{tab:architecture}
\end{table*}

\section{Background}
\paragraph{Transfer learning}
The goal of transfer learning is to pre-train a model on one dataset so that it performs well with little extra training on datasets with few labeled examples.
Specifically with bio-signals, these downstream datasets are time series data, often with a unique temporal length, sampling frequency, and a variety of different prediction targets.
Labels in the datasets are provided for windows over time, e.g. when is a cardiac rhythm normal and when is it indicative of atrial fibrillation.
A common self-supervised pre-training method is masked auto-encoding~\cite{he2022masked}, which has also been used for bio-signal transfer learning previously~\cite{liu2023frequency,dong2024simmtm}.
Specifically, masked autoencoding models can be loosely split into two parts; a patching and encoding part (encoder), and a transformer, see Figure~\ref{fig:model}a.
In the simplest case, the patching and encoding part of the model linearly embeds non-overlapping windows (patches) of the input time-series into encoded features, see Figure~\ref{fig:model}a.
A random subset of the encoded features are then masked by replacing them with a mask token, and used as input to the transformer, which produces context features.
These context features are passed through a pretrain head to reconstruct the original signal corresponding to the patches that are masked out. The pretrain head can be any type of neural network architecture, for example a linear layer or a multi-layer perceptron (MLP).
Masked auto-encoding thus forces the transformer to learn temporal relationships between patches so it can predict the masked patches.

After pre-training the model, the pretrain head is removed from the model, and in the fine-tuning stage, a linear layer is attached either to the encoded features (after the encoder) or the context features (after the transformer), see Figure~\ref{fig:model}a.
The linear layer, together with the rest of the model, is trained on the downstream bio-signal dataset.
The assumption behind this method is that the relevant features and patch relationships the transformer and/or the encoder have learned are transferable to new bio-signal datasets and are a better starting point for supervised training than a randomly initialized network.

\paragraph{Related work}
In the supervised and decoding literature for EEG data, convolution-transformer hybrids have shown improvements over convolution-only architectures on a variety of tasks~\cite{song2022eeg,peh2022transformer,gong2023eeg,miltiadous2023dice}. However, all of these methods are specifically developed for EEG, are fully supervised, and have spatial convolution blocks that make it hard for them to transfer to new data with a different number of channels. Moreover, the field of masked auto-encoding for EEG and bio-signal data has recently seen many advances in same-data pre-training and fine-tuning~\cite{chien2022maeeg,yang2024biot}, representation learning~\cite{foumani2024eeg2rep}, semi-supervised learning~\cite{eldele2023self}, and cross-modal learning~\cite{deldari2023latent}. These models do not perform pre-training on one bio-signal with fine-tuning on a new bio-signal, however.

One reason why transfer learning from one bio-signal dataset to another dataset likely works is because important frequency ranges and thus low-level statistics used for prediction are similar across bio-signals~\cite{neyshabur2020being}.
Since and time-frequency features are important for the analysis or predictions from bio-signals such as EEG~\cite{durongbhan2019dementia}, ECG~\cite{odinaka2010ecg}, and EMG~\cite{weiderpass2013time}, learning features that are consistent across time and frequency space can be helpful for downstream predictions.
Time-frequency consistency (TFC)~\cite{zhang2022self} presents a method to extract features from the time and frequency domains that are trained to be similar. To do this, they use contrastive learning, and introduce new frequency-based augmentations. Alternatively, bioFAME~\cite{liu2023frequency}, introduces a Fourier neural operator (FNO)-based encoder~\cite{li2020fourier,guibas2021adaptive} to directly learn features from the frequency space. Additionally, they use a frequency-aware masked autoencoder for multimodal pre-training, and introduce PatchTST~\cite{nie2022time} as a baseline. 
Recently, SimMTM~\cite{dong2024simmtm} has improved on TFC with the same model architecture by relating masked auto-encoding to manifold learning. They introduce a new pre-training task where the model needs to reconstruct the original timeseries with a set of masked timeseries outside the manifold.
These previous works motivated us to develop a convolutional-transformer hybrid model (CiTrus), with two pre-training techniques, and a better way of transfering a model from one bio-signal dataset to another.

\begin{table*}[!t]
\centering
\begin{tabular}{l|ccc|ccc|ccc|ccc|}
\hline
Dataset & \multicolumn{3}{c}{EMG} & \multicolumn{3}{c}{ECG} & \multicolumn{3}{c}{PPG} & \multicolumn{3}{c}{HAR} \\
\hline
Data percentage & 20\% & 50\% & 80\% & 0.5\% & 1\% & 2\% & 10\% & 20\% & 50\% & 10\% & 20\% & 50\% \\
\hline
PatchTST & \textbf{+18.7} & -12.4 & -11.1 & +2.8 & \textbf{+3.5} & +1.9 & +1.1 & \textbf{+1.5} & +1.3 & -7.7 & -6.2 & \textbf{-2.5} \\
bioFAME & \textbf{+49.0} & +27.7 & +10.0 & +5.0 & +4.4 & \textbf{+5.2} & \textbf{+3.3} & -0.6 & +2.0 & +10.5 & +7.9 & \textbf{+15.3} \\
NLPatchTST (p) & \textbf{+12.5} & +1.9 & +0.1 & \textbf{+3.4} & -0.1 & -0.1 & \textbf{+6.4} & +2.5 & -0.9 & -2.9 & -3.7 & \textbf{-0.8} \\
NLPatchTST (mp) & \textbf{+16.7} & +1.0 & -1.4 & \textbf{+5.6} & +1.5 & +1.1 & \textbf{+5.2} & +1.8 & +0.3 & -7.8 & -7.4 & \textbf{-1.7} \\
SimMTM & -16.1 & -3.2 & \textbf{-0.2} & \textbf{+3.5} & -7.5 & -3.5 & -4.0 & \textbf{+0.2} & -1.6 & \textbf{+2.0} & +0.6 & +0.5 \\
Ci & -1.1 & -0.3 & \textbf{-0.2} & -0.8 & -1.4 & \textbf{+1.9} & \textbf{+1.9} & -0.2 & +0.3 & -8.5 & -3.8 & \textbf{-0.3} \\
CiTrus (p) & -4.4 & -0.7 & \textbf{-0.2} & -3.5 & -1.2 & \textbf{+0.1} & \textbf{+6.5} & +2.6 & -0.3 & +2.0 & \textbf{+2.3} & -0.6 \\
CiTrus (fp) & \textbf{+0.4} & -0.9 & +0.3 & +8.3 & \textbf{+8.9} & +7.2 & \textbf{+12.9} & +12.7 & +9.0 & -6.5 & -4.2 & \textbf{-0.3} \\
CiTrus (mp) & -2.0 & -2.0 & \textbf{-0.2} & -0.8 & +0.5 & \textbf{+1.8} & \textbf{+7.1} & +3.3 & -1.4 & +1.0 & \textbf{+2.4} & -1.3 \\
\hline
Dataset & \multicolumn{3}{c}{FDB} & \multicolumn{3}{c}{Epilepsy} & \multicolumn{3}{c}{Gesture} & \multicolumn{3}{c}{SleepEDF} \\
\hline
Data percentage & 0.5\% & 1\% & 2\% & 1\% & 5\% & 10\% & 10\% & 20\% & 50\% & 0.5\% & 1\% & 2\% \\
\hline
PatchTST & \textbf{+7.0} & +6.9 & -1.8 & \textbf{+0.8} & -0.1 & +0.0 & \textbf{+12.2} & +8.7 & +4.8 & \textbf{+7.3} & +0.3 & +1.3 \\
bioFAME & +7.3 & \textbf{+8.9} & +8.4 & \textbf{+0.5} & +0.0 & +0.1 & \textbf{+16.8} & +11.8 & +10.0 & +0.4 & \textbf{+0.6} & -1.2 \\
NLPatchTST (p) & \textbf{+4.1} & +1.7 & -3.1 & +13.4 & \textbf{+15.0} & +4.5 & \textbf{+11.4} & +7.3 & +4.4 & \textbf{+13.7} & +4.4 & +7.1 \\
NLPatchTST (mp) & \textbf{+6.1} & +3.6 & -1.1 & +13.4 & \textbf{+15.6} & +4.4 & \textbf{+15.1} & +9.1 & +5.4 & \textbf{+7.2} & +1.8 & +4.3 \\
SimMTM & \textbf{+33.5} & +19.2 & +5.1 & -0.3 & -0.3 & \textbf{-0.0} & \textbf{+3.0} & +1.2 & +0.5 &  &  & \\
Ci & -1.9 & -1.8 & \textbf{-0.7} & \textbf{+0.3} & +0.3 & +0.2 & \textbf{+13.7} & -6.5 & -0.5 & +0.3 & +0.0 & \textbf{+1.7} \\
CiTrus (p) & \textbf{+2.1} & -3.3 & -2.2 & -0.9 & -0.8 & \textbf{-0.1} & -1.3 & -6.1 & \textbf{+0.4} & \textbf{+5.1} & -1.0 & +1.5 \\
CiTrus (fp) & \textbf{+8.8} & +1.1 & +1.3 & -0.8 & \textbf{+0.1} & -0.0 & \textbf{+11.8} & +1.2 & +5.6 & -22.1 & -18.9 & \textbf{-9.3} \\
CiTrus (mp) & \textbf{+1.1} & -4.8 & -2.1 & -1.2 & -0.4 & \textbf{-0.3} & \textbf{+9.1} & +3.0 & -0.4 & \textbf{+0.1} & -4.0 & -1.8 \\
 
\hline
\end{tabular}
\caption{A comparison between a pre-trained and fine-tuned version of the model, and the same model trained from scratch on the downstream datasets; (p) means it is pre-trained, (mp) means it uses multi-modal pre-training, and (fp) means it uses frequency pre-training. Each value is the average percentage improvement (across all three metrics) of pre-training compared to training from scratch. Values are made bold for the data regime within a dataset where pre-training and fine-tuning most increases the performance to show the effect of data availability on pre-training improvements.}
\label{tab:pretraining}
\end{table*}

\section{Method}
\paragraph{CiTrus: Convolution transformer hybrid}
Given the potential importance of the frequency representation in bio-signal transfer learning, it is a natural choice to look at convolutions.
Convolutional networks have learnable filters that can learn representations in the frequency domain, and are good at capturing local features in the time series.
We propose a model that combines the best of both worlds that we call CiTrus, a [C]onvolution-[Tr]ansformer hybrid, as shown in Figure~\ref{fig:model}b.
The encoder in our model is a $3$-layer residual convolutional network (a more detailed description of the convolutional network can be found in Appendix~\ref{app:model}), and the transformer is a PatchTST~\cite{nie2022time} transformer.
Our reason for choosing a PatchTST transformer, similar to bioFAME, is that it exhibits especially good performance for masked auto-encoding and is channel-independent.
To create channel-independence for the convolutional encoder, we concatenate the number of channels along the batch, and both pre-train and fine-tune the model with a single channel as input. This both increases the parameter efficiency and increases the number of samples the weights in the convolutional encoder are trained with, but also allows the model to easily transfer to data with a different number of channels. Specifically, without making the convolutional encoder channel-independent we can’t transfer a model trained with 3 channels to a downstream dataset with a single channel.

\begin{table*}[!t]
\centering
\begin{tabular}{l|ccc|ccc|ccc|ccc|}
\hline 
 Dataset & \multicolumn{3}{c}{EMG} & \multicolumn{3}{c}{ECG} & \multicolumn{3}{c}{PPG} & \multicolumn{3}{c}{HAR} \\ 
 \hline 
 Data percentage & 20\% & 50\% & 80\% & 0.5\% & 1\% & 2\% & 10\% & 20\% & 50\% & 10\% & 20\% & 50\% \\ 
 \hline 
NLPatchTST & \textbf{+17.3} & \textbf{+7.5} & \textbf{+6.0} & \textbf{+2.2} & \textbf{+3.6} & \textbf{+1.9} & -0.5 & -0.3 & \textbf{+1.6} & -4.9 & -3.7 & -0.9 \\ 
CiTrus & \textbf{+6.3} & \textbf{+8.1} & \textbf{+3.1} & \textbf{+6.0} & \textbf{+3.8} & \textbf{+2.4} & \textbf{+1.1} & \textbf{+1.5} & -0.6 & -0.8 & \textbf{+0.2} & -0.7 \\ 
\hline 
 Dataset & \multicolumn{3}{c}{FDB} & \multicolumn{3}{c}{Epilepsy} & \multicolumn{3}{c}{Gesture} & \multicolumn{3}{c}{SleepEDF} \\ 
 \hline 
 Data percentage & 0.5\% & 1\% & 2\% & 1\% & 5\% & 10\% & 10\% & 20\% & 50\% & 0.5\% & 1\% & 2\% \\ 
 \hline 
NLPatchTST & -2.0 & \textbf{+0.1} & -0.9 & \textbf{+0.1} & \textbf{+0.5} & -0.1 & \textbf{+3.8} & \textbf{+1.9} & \textbf{+1.3} & -5.5 & -2.5 & -2.5 \\ 
CiTrus & -0.5 & \textbf{+0.5} & -0.1 & -0.2 & \textbf{+0.4} & -0.1 & \textbf{+12.7} & \textbf{+11.1} & -0.3 & -4.7 & -3.0 & -3.2 \\ 
\hline
\end{tabular}
\caption{A comparison of multimodal pre-training to unimodal pre-training, where each value is the average improvement (across all three metrics) in percentages. Values that are larger than $0$, indicating performance improvement, are made bold.}
\label{tab:multimodal}
\end{table*}

\paragraph{Pre-training (p)}
In conventional masked-autoencoding (MAE), the timeseries first needs to be segmented into non-overlapping patches of equal length. Formally, let $x \in \mathbb{R}^{T}$ be a bio-signal with T timesteps, a patch size $S$, and assume $T\%S = 0$. The timeseries is then split into T/S = P patches, to obtain a signal $x_p \in \mathbb{R}^{P\times S}$. The patched signal is masked by replacing $X\%$ of the patches with a mask token and a transformer is trained to reconstruct the masked patches using the information in the non-masked patches. Recent work has shown that it is beneficial to encode the signal into a latent space before patching and masking. Therefore, in our model, we propose to use a CNN as the encoder before patching. Formally, let $x \in \mathbb{R}^{1\times T}$, then CNN(x) $\in \mathbb{R}^{ D\times P}$, with D the number of output channels, and P the number of patches. The stride of the CNN determines the stride of the patches, and the receptive field determines the “effective” patch length. Since the receptive fields for each patch overlap in the data space, typical masking won’t work (neighboring patches have too much information about each other). The inputs to the transformer are thus P patches with D dimensions, masked with our proposed masking method for hybrid models. The pre-train head used for this type of pre-training is a flipped version of the convolutional encoder. In order to ablate the impact of the convolutional encoder, we develop a new baseline, by using a $3$-layer multi-layer perceptron (MLP) to encode and decode each patch. We refer to this model as NLPatchTST (Non-Linear PatchTST) to differentiate from the original PatchTST model that uses a linear embedding layer.

\paragraph{Frequency pre-training (fp)}
Besides predicting the original input time-series signal, we can also push the model to explicitly predict frequency representations of the original signal by predicting the spectrogram that corresponds to the masked patches instead of the original signal. The same low-frequencies can be repeated across patches, so to make the prediction of the spectrogram more challenging, we z-score the spectrograms along the time dimension. This new pre-training method is visualized in Figure~\ref{fig:model}b. Its exact settings and implementation are discussed in Appendix~\ref{app:frequency-pretraining}. 

\paragraph{Multimodal pre-training}
Given that the utility of multimodal pre-training was verified for the bioFAME model, we explore how well multimodal pre-training works for CiTrus. To pre-train CiTrus with multimodal data, each modality is passed through the convolutional encoder and transformer independently. This ensures that the convolutional encoder and transformer see data from both modalities and can learn to adapt its weights to both. In our work we follow bioFAME and use two modalities, EEG and EOG, during pre-training, with a separate decoder for each modality. Specifically, the EEG and EOG signals are concatenated along the batch dimension in the convolutional encoder and transformer. After the transformer, the EEG and EOG context features are separated, and concatenated along the latent dimension, see Figure~\ref{fig:model}c. The context features are then used as input for each decoder to predict the masked patches in each modality. To encourage cross-modality learning, we mask patches independently for each modality, allowing information for a masked patch in one modality to be available from the other modality.

\paragraph{Transfer learning to new datasets}
Previous works, such as TFC and bioFAME transfer a pre-trained model to new downstream datasets with different temporal lengths by interpolating the data to match the signal length of the pre-training dataset.
This causes a mismatch in sampling frequencies between the downstream fine-tuning datasets and the pre-train dataset. The potential frequency representation learned by the encoder and the temporal relationships learned by the transformer on the pre-training dataset are now potentially irrelevant because they act on a different frequency range and timescale.
We therefore propose a new strategy where the fine-tuning dataset is resampled to match the frequency of the pre-training dataset.
This allows the model to attend to similar frequency spectra in the fine-tuning data as the pre-training data.
If the length of the resampled fine-tuning data is now longer than the temporal length of the pre-training signal, we use an overlapping sliding window approach, and average the embeddings across the windows.
If the length of the re-sampled fine-tuning data is shorter, we pad the signal with zeros.

\begin{table*}[!t]
\centering
\begin{tabular}{l|ccc|ccc|ccc|ccc|}
\hline 
 Dataset & \multicolumn{3}{c}{ECG} & \multicolumn{3}{c}{PPG} & \multicolumn{3}{c}{HAR} & \multicolumn{3}{c}{Gesture} \\ 
 \hline 
 Data percentage & 0.5\% & 1\% & 2\% & 10\% & 20\% & 50\% & 10\% & 20\% & 50\% & 10\% & 20\% & 50\% \\ 
 \hline 
PatchTST (s) & \textbf{+4.5} & -0.3 & -0.7 & \textbf{+58.9} & \textbf{+60.0} & \textbf{+58.9} & \textbf{+23.9} & \textbf{+22.1} & \textbf{+19.7} & \textbf{+16.0} & \textbf{+18.5} & \textbf{+9.5} \\ 
PatchTST (p) & \textbf{+6.7} & \textbf{+10.1} & \textbf{+3.5} & \textbf{+52.9} & \textbf{+58.9} & \textbf{+57.8} & \textbf{+14.8} & \textbf{+17.2} & \textbf{+18.8} & \textbf{+3.7} & \textbf{+5.6} & \textbf{+4.2} \\ 
bioFAME (s) & \textbf{+5.2} & \textbf{+2.9} & \textbf{+2.6} & \textbf{+60.7} & \textbf{+55.9} & \textbf{+47.7} & \textbf{+16.7} & \textbf{+15.3} & \textbf{+16.0} & \textbf{+7.8} & \textbf{+10.9} & \textbf{+9.5} \\ 
bioFAME (mp) & \textbf{+15.4} & \textbf{+11.4} & \textbf{+9.8} & \textbf{+64.1} & \textbf{+47.8} & \textbf{+56.2} & \textbf{+14.6} & \textbf{+16.6} & \textbf{+26.5} & -0.3 & \textbf{+2.1} & \textbf{+0.1} \\ 
SimMTM (s) & \textbf{+4.4} & \textbf{+15.3} & \textbf{+8.1} & \textbf{+30.1} & \textbf{+29.9} & \textbf{+51.9} & \textbf{+0.2} & -0.1 & -0.4 & -3.8 & \textbf{+0.0} & -1.7 \\ 
SimMTM (p) & \textbf{+1.4} & \textbf{+1.7} & -1.8 & \textbf{+24.3} & \textbf{+26.9} & \textbf{+28.4} & -0.3 & -0.6 & -0.1 & -0.1 & \textbf{+0.6} & -0.7 \\ 
NLPatchTST (s) & -0.5 & -2.0 & -2.6 & \textbf{+49.6} & \textbf{+50.4} & \textbf{+64.3} & \textbf{+20.9} & \textbf{+20.7} & \textbf{+15.8} & \textbf{+8.8} & \textbf{+14.2} & \textbf{+3.0} \\ 
NLPatchTST (p) & \textbf{+3.3} & \textbf{+3.5} & -0.5 & \textbf{+58.6} & \textbf{+55.9} & \textbf{+61.8} & \textbf{+17.1} & \textbf{+20.1} & \textbf{+18.6} & -0.6 & \textbf{+3.8} & \textbf{+3.5} \\ 
NLPatchTST (mp) & \textbf{+5.5} & \textbf{+8.0} & -0.2 & \textbf{+57.6} & \textbf{+52.5} & \textbf{+64.9} & \textbf{+10.6} & \textbf{+14.7} & \textbf{+18.8} & \textbf{+3.3} & \textbf{+6.4} & \textbf{+4.9} \\ 
Ci (s) & \textbf{+0.2} & \textbf{+3.8} & \textbf{+2.6} & \textbf{+29.1} & \textbf{+30.4} & \textbf{+20.4} & \textbf{+22.7} & \textbf{+17.8} & \textbf{+13.2} & \textbf{+8.2} & \textbf{+4.9} & \textbf{+2.7} \\ 
Ci (p) & -1.2 & \textbf{+1.6} & \textbf{+2.9} & \textbf{+39.1} & \textbf{+41.7} & \textbf{+31.9} & \textbf{+18.5} & \textbf{+20.4} & \textbf{+11.7} & \textbf{+0.8} & \textbf{+3.9} & \textbf{+4.1} \\ 
CiTrus (s) & \textbf{+0.4} & \textbf{+6.0} & \textbf{+1.0} & \textbf{+32.1} & \textbf{+24.8} & \textbf{+21.1} & \textbf{+16.7} & \textbf{+18.7} & \textbf{+17.6} & \textbf{+21.1} & \textbf{+8.8} & \textbf{+3.7} \\ 
CiTrus (p) & \textbf{+6.7} & \textbf{+6.7} & \textbf{+8.6} & \textbf{+49.4} & \textbf{+51.0} & \textbf{+47.3} & \textbf{+19.3} & \textbf{+17.7} & \textbf{+14.3} & -0.9 & \textbf{+0.5} & \textbf{+2.6} \\ 
CiTrus (fp) & \textbf{+0.5} & \textbf{+4.5} & \textbf{+3.1} & \textbf{+50.3} & \textbf{+48.1} & \textbf{+30.8} & \textbf{+11.5} & \textbf{+14.8} & \textbf{+17.7} & \textbf{+9.0} & \textbf{+2.8} & \textbf{+3.4} \\ 
CiTrus (mp) & \textbf{+11.0} & \textbf{+5.9} & \textbf{+8.0} & \textbf{+51.6} & \textbf{+49.7} & \textbf{+44.2} & \textbf{+17.4} & \textbf{+16.9} & \textbf{+14.2} & \textbf{+0.8} & \textbf{+6.9} & \textbf{+1.7} \\ 

\hline
\end{tabular}
\caption{A comparison between the previously used fine-tuning technique (temporal interpolation), and our proposed fine-tuning technique; (s) is trained from scratch, (p) means it is pre-trained, (mp) means it uses multi-modal pre-training, and (fp) means it uses frequency pre-training. Each value indicates the average performance improvement (across all three metrics) in percentages. Values larger than $0$ are made bold.}
\label{tab:sliding-window}
\end{table*}

\section{Experiments}

\paragraph{Evaluation framework \& datasets}
Similar to previous work~\cite{zhang2022self}, we use the SleepEDF, a large sleep EEG dataset~\cite{kemp2000analysis} for pre-training. The dataset consists of $30$ second windows of EEG data sampled at $100$Hz. For the uni-modal pre-training we use the two EEG channels that are captured. For the multi-modal pre-training we use the two EEG channels and the (single-channel) EOG data that is from the same dataset.
As downstream datasets, we use $4$ bio-signal datasets that were used to evaluate the TFC, bioFAME, and SimMTM models with: 
\begin{itemize}
\item An electromyography (EMG) dataset~\cite{goldberger2000physiobank} with a 375ms window length, sampled at 4KHz, and a single channel.
\item A gesture recognition dataset~\cite{liu2009uwave} with a 3.15s window length, sampled at 100Hz, and 3 channels.
\item An EEG Epilepsy dataset with a 1.02s window size, sampled at 174Hz, and a single channel.
\item An electromotor fault-detection (FD-B)~\cite{lessmeier2016condition} dataset with a 80ms window size, sampled at 64KHz, and a single channel.
\end{itemize}
Additionally, to increase the diversity of downstream tasks and modalities we test, we also add additional datasets:
\begin{itemize}
\item A photoplethysmography (PPG) dataset~\cite{schmidt2018introducing} with a 60s window length, sampled at 64Hz, and a single channel.
\item The HAR dataset~\cite{reyes2015smartphone} with a 2.56s window length, sampled at 50Hz, and 6 channels.
\item An electrocardiogram (ECG) dataset~\cite{moody1983new} with a 10s window length, sampled at 250Hz, and 2 channels.
\item The SleepEDF test set 
\end{itemize}
More information about these datasets can be found in Appendix~\ref{app:data}.

The training, validation, and testing data splits were initially defined in the TFC work for the EMG, Gesture, FD-B, and Epilepsy datasets, and these splits were also used in bioFAME and SimMTM.
Although we evaluate our proposed model on those previously defined data splits, see Appendix~\ref{app:original-data}, we find that the variance in performance across test folds is much larger than across random seeds, see Table~\ref{tab:variance} in Appendix~\ref{app:variance}.
Thus, to more comprehensively evaluate models used for transfer learning, we advocate for and implement a cross-validation procedure that averages test performance across $10$-fold test splits.
The training and validation data (the remaining $9$ folds) is then reduced to the specific data-regime percentage.
This leftover percentage of training and validation data is then split into $75\%$ training and $25\%$ validation data.
Averaging the performance of the models across random seeds and test splits helps average out some of the performance's randomness.
We discuss the exact protocol we use to create the test splits in Appendix~\ref{app:kfold}.
For all the downstream tasks (except the SleepEDF test set), we follow the TFC work and split the data into $2$ second windows before pre-training because the downstream data is often very short.
For the SleepEDF evaluation, we keep the original $30$s for pre-training.
We also compare $30$ vs $2$ second window pre-training in Appendix~\ref{app:time-length}.
To understand the relationship between transfer learning and the amount of data available in the downstream datasets during we evaluate each model across a range of data percentages during fine-tuning.

\paragraph{Experimental settings}
The implementations for the bioFAME, SimMTM, and PatchTST models are taken from their respective official implementations. The following are the model hyperparameters for all the models, except SimMTM, for which we use the official implementation's hyperparameters. The hyperparameters are matched to the bioFAME paper as much as possible to make the comparisons as fair as possible. We use a $4$-layer transformer (with $64$ latent dimensions, $128$ feed-forward dimension and $8$ heads), a $3$-layer convolutional network with $32$ channels in the first residual convolution layer that double every layer (only applicable for CiTrus), a patch size of $20$, a $0.5$ masking ratio, and a block masking size of $5$ (only applicable for CiTrus). All models are pre-trained for $200$ epochs with a $128$ batch size, and a $0.0001$ learning rate. All models are fine-tuned for $100$ epochs with a $64$ batch size, without any augmentations, and the same learning rate as during pre-training. We use the last pre-training model checkpoint for fine-tuning, and evaluate the best (based on the validation set) fine-tuning checkpoint on the test set. Our models are pre-trained and fine-tuned on an AWS instance with $4$ NVIDIA A10 GPUs, with $[42, 1337, 1212, 9999]$ as the model seeds, and $42$ as the seed for data randomization and fold generation. Given the $10$ data folds, there are $40$ runs per model, per data regime and per dataset. For a deeper explanation of the model settings, see Appendix~\ref{app:model}. Since the evaluation is done on classification tasks, we report the accuracy (ACC), area under the curve of the receiver operating characteristic (ROC), and area under the precision-recall curve (PRC) to get a variety of classification metrics. Lastly,
standard deviations and Wilcoxon signed-rank test outcomes for each experiment are discussed in Appendix~\ref{app:experimental-statistics}.

\paragraph{Model architecture comparisons}
The PatchTST, bioFAME, SimMTM are evaluated and compared to our NLPatchTST and CiTrus models for the transfer-learning task in Table~\ref{tab:architecture}. To understand the effect the transformer in our model has on the performance, we also compare Ci, which uses the encoded features, to CiTrus, which uses the context features. For each dataset, except the EMG and FD-B datasets, we use our proposed sliding window approach during fine-tuning. The EMG and FD-B datasets have windows for predictions that are too short for our sliding window technique, so we interpolate the data to $200$ timesteps as described in the TFC work. Moreover, to limit the size of the table, we report the hardest data regime; the lowest amount of data available for each dataset. The percentage of training + validation data that is available is mentioned after the name of the dataset in Table~\ref{tab:architecture}. Note, since the convolutional encoder used in the SimMTM model only works with inputs that are $200$ samples in length, it is not compatible with the SleepEDF test set, so we do not report the result.

The results in Table~\ref{tab:architecture} indicate that convolution-based models (SimMTM, Ci, and CiTrus) almost always achieve the best performance. For ECG, PPG, and FD-B, using the context features (CiTrus)  performs better than using the encoded features (Ci). The CiTrus model with pre-training and fine-tuning performs the best on those datasets as well. For every dataset, except the Gesture dataset, the CiTrus model is atleast in the top three best models. The fact that both the HAR and Gesture datasets are accelerometer-based datasets could explain the lower performance on these datasets. Our frequency-based pre-training model performs the best on average, see Appendix~\ref{app:avg-results}.

\paragraph{Training from scratch vs pre-training}
\label{sec:pretraining}
To understand the effect of pre-training on downstream performance, we calculate how much better each model performs after pre-training. Since we use three metrics in Table~\ref{tab:architecture} to evaluate each approach, we calculate the average percentage pre-train improvement across the three metrics over training the model from scratch. We also bold the largest improvement across the data regimes for each model and each dataset in Table~\ref{tab:pretraining}.

For most settings in Table~\ref{tab:pretraining} pre-training improves downstream performance, but this improvement depends on the dataset and the underlying structure of the model. We can see dramatic increases in performance for transformer-based models across all data regimes on the EMG dataset. However, gains for the Ci model, which is fully convolutional, are marginal. Moreover, pre-training improves performance the most for the lowest data regime on the Gesture, PPG, and SleepEDF datasets.

\paragraph{Multimodal vs unimodal pre-training}
For both the CiTrus and the NLPatchTST model we investigate whether multimodal pre-training further improves performance over unimodal pre-training.
Similar to the pre-training results, we compute the average percentage improvement across the three metrics with respect to uni-modal pre-training. Results that improve performance over uni-modal pre-training are made bold in Table~\ref{tab:multimodal}.

There are a few datasets where multimodal pre-training almost always performs better: the EMG, ECG, and Gesture datasets. These show clear improvements for multimodal pre-training. Moreover, the CiTrus model generally benefits more from multi-modal pre-training than the NLPatchTST model. This could be due to the inherent ability of convolutional networks to learn local feature representations jointly across multiple modalities. Interestingly, for the SleepEDF test set, multimodal pre-training degrades performance, potentially because only EEG data is available during fine-tuning.

\subsection{Resampling-adaptive fine-tuning}
To verify how well our proposed fine-tuning approach works for new datasets with different temporal lengths, sampling frequencies, and modalities, we compare it to the interpolation method used in the TFC, bioFAME, and SimMTM works. For each model, we compute the average percentage improvement with our proposed fine-tuning technique over fine-tuning with temporal interpolation. For this experiment we select datasets that have very different sampling frequencies than the original pre-training dataset. We exclude the EMG and FD-B datasets, since their lengths are extremely short ($<400ms$), and lead to very small temporal samples ($<40$) at $100$Hz. 

In Table~\ref{tab:sliding-window} the average percentage increase in metrics shows that in almost all cases our new fine-tuning approach markedly improves performance. Especially on the PPG dataset the new approach reaches improvements of 64\%. Moreover, improvements are achieved for both convolutional and transformer-based models. Pre-trained models benefit the most from the new fine-tuning approach. 

\section{Discussion}
In this work, we assessed the transferability of neural networks pre-trained on a large EEG dataset and fine-tuned on low-data downstream bio-signal datasets. First, we generally find that models with convolutions (SimMTM, Ci, and CiTrus) perform the best. We suspect this is largely due to their inductive bias and parameter efficiency. Convolutions share many of their weights when processing the timeseries, and given that convolutions are learnable frequency filters, they transfer well to datasets where features come from similar parts of the frequency spectrum. Second, the idea that the parameter efficiency and inductive bias of convolutions is key in low-data settings is further supported by the fact that transformer-based models improve the most with pre-training, especially in low-data settings, which indicates that they require more data to learn the important temporal relationships required for predictions. This is also true for other types of data, such as images, where the vision transformer (ViT)~\cite{dosovitskiy2020image}  starts outperforming convolutions when pre-trained with a large dataset. Third, multimodal pre-training often improves downstream fine-tuning performance even more. Fourth, we find that pre-training is not necessary for the EMG and FD-B datasets. Although the TFC, bioFAME, and SimMTM models found models pre-trained on bio-signal data to transfer well to FD-B, which is an electromotor dataset, we find that our Ci model trained from scratch outperforms all other models. Additionally, datasets where performance is already high, like the Epilepsy dataset, do not benefit much from pre-training either. Fifth, CiTrus with frequency-based pre-training performs the best on average across datasets for the lowest and highest data regimes we tested. Sixth, we find that variance across test folds is much higher (up to $4609\%$) than across random seeds and thus strongly recommend evaluating transfer learning models across different test sets to more robustly evaluate their performance. Lastly, our fine-tuning approach is better in essentially all cases for all models, and improves performance up to $60\%$. Our approach improves performance the most for pre-trained models, which we believe is because the temporal relationships the encoder and transformer learn are now aligned between the pre-training and downstream datasets.

\clearpage
\bibliography{aaai25}

\clearpage
\onecolumn
\appendix
\section{K-fold generation}
\label{app:kfold}
The K-fold generation is split into two strategies, based on how the original test split was defined.
For the EMG, FD-B, Epilepsy, and Gesture datasets, which were used in~\cite{zhang2022self}, the data was concatenated.
Then, a StratifiedKfold object was instantiated~\cite{sklearn_api}, with $10$ folds, shuffling, and its random state set to $42$.
This splits the data into $10$ folds, and we take each split as a test fold once.
The remaining $9$ folds are used as a training + validation set.
Then, depending on the data regime, we use stratified train\_test\_split() to only keep X\% of the training + validation data, with random\_state=42.
This is to simulate specific data regimes, X here refers to the percentages used in the main text and the rest of the Appendix.
Lastly, we perform a stratified split to split the leftover training + validation data into a training and validation set, with $75\%$ of the remaining data in the training set, and random\_state=42.
The only difference in this strategy for the other datasets: HAR, ECG, and PPG, is that they are split into a training, validation, and test set based on the subject the windows come from.
Thus, to ensure the folds reflect subjects instead of just doing a stratified split across the windows.
For the ECG and PPG dataset, we instantiate a KFold object with $10$ folds, and for HAR (because there are multiple timeseries per subject) we instantiate a GroupKfold object.
Then, for the ECG and PPG datasets, the subjects are split into $10$ folds, for the HAR dataset, the timeseries are split, based on the subjects as groups, into $10$ splits.
The rest of the procedure is the same as before, except that stratification in case of leaving X\% of the data is done based on both subject and label, i.e. we keep each class and subject in the leftover training + validation dataset atleast once.
For the SleepEDF dataset we do not create $10$-fold splits to fine-tune on because of time limits.
Namely, each split would require a complete pre-training cycle, which are costly in terms of time.
Thus, the results for the SleepEDF fine-tuning are averaged only over the random seeds.

\clearpage
\section{Results on pre-defined splits from TFC paper}
\label{app:original-data}
The following table, Table~\ref{tab:original-split}, shows the performance of the models on the original (single) test split. The models are pre-trained and fine-tuned in the way described in our main text. Since we find that resampling to $30$s after pre-training on $30$s windows performs the best for the EMG and FD-B datasets, when we use a resampling/interpolation technique, see Appendix~\ref{app:time-length}, we use the models pre-trained on $30$s windows in the SleepEDF dataset, similar to the BioFAME paper.
Then, the fine-tuning dataset is resampled to $30$s (3000 timesteps) based on the method used in~\cite{liu2023frequency}, linear interpolation.
This method only deviates for the SimMTM and CiTrus (fp) models because the SimMTM model does not accept $30$s inputs, and the CiTrus (fp) model performs better with $2$s inputs.
SimMTM~\cite{dong2024simmtm} follows the original TFC paper~\cite{zhang2022self} by pre-training on ~$2$s splits, and the SimMTM (s) is the same model architecture as TFC trained from scratch.
Lastly, for the Gesture and Epilepsy datasets, the data is simply padded to $200$ timesteps (from 178) for the Epilepsy dataset, and resampled (from 206) to $200$ for the Gesture dataset, and the models pre-trained on $2$s (200 timesteps) splits of the SleepEDF dataset are used.
This is done because both dataset's temporal length is almost the same as the $2$s pre-training windows (200 timesteps).
For the BioFAME model, we find the same performance within the standard deviation reported in the original paper for the EMG dataset. However, for the FD-B and Epilepsy datasets the performance is slightly lower than the original paper. This may be because we do not use any label augmentations, such as MixUp~\cite{zhang2017mixup} during the fine-tuning stage. 
For the SimMTM paper we find the same performance, except on the EMG dataset.
Generally, we tried to match our framework as closely as possible to bioFAME and SimMTM, but since they each use different pre-training lengths, it can be hard to exactly replicate the reported performances.
\FloatBarrier
\begin{table*}[ht]
\begin{tabular}{l|ccc|ccc|ccc|ccc|}
\hline 
 Dataset & \multicolumn{3}{c}{EMG} & \multicolumn{3}{c}{FD-B} & \multicolumn{3}{c}{Epilepsy} & \multicolumn{3}{c}{Gesture} \\ 
 \hline 
 Metric & ACC & ROC & PRC & ACC & ROC & PRC & ACC & ROC & PRC & ACC & ROC & PRC \\ 
 \hline 
PatchTST (s) & 95.12 & 98.06 & 98.79 & 69.16 & 57.41 & 43.57 & 93.89 & 50.25 & 80.35 & 61.67 & 71.18 & 40.96 \\ 
PatchTST (p) & 94.51 & \doubleunderline{99.54} & \underline{99.15} & 68.8 & 57.53 & 43.51 & \underline{95.47} & 50.24 & 80.35 & 62.5 & 70.42 & 40.56 \\ 
bioFAME (s) & \underline{95.73} & \underline{99.33} & 97.14 & 71.48 & 57.96 & 44.19 & 89.62 & 50.24 & 80.37 & 46.46 & 70.41 & 38.7 \\ 
bioFAME (mp) & \textbf{98.17} & \textbf{99.88} & \textbf{99.85} & 70.4 & 57.77 & 43.93 & 93.76 & 50.24 & 80.36 & 55.42 & 71.84 & 40.47 \\ 
NLPatchTST (s) & \underline{95.73} & 97.88 & 98.67 & 74.69 & 58.54 & 44.72 & 93.05 & 50.26 & 80.37 & 62.71 & 72.56 & 41.65 \\ 
NLPatchTST (p) & 95.12 & 98.96 & \doubleunderline{99.36} & 79.08 & 59.18 & 45.48 & 95.11 & 50.25 & 80.36 & 63.96 & \textbf{73.27} & 42.55 \\ 
NLPatchTST (mp) & 95.12 & 98.33 & 98.92 & 69.73 & 57.84 & 43.87 & \doubleunderline{95.51} & 50.25 & 80.37 & 63.54 & 72.4 & 41.8 \\ 
SimMTM (s) & 92.68 & 97.24 & 91.11 & 62.66 & 57.03 & 41.14 & 93.2 & 50.25 & 80.37 & \doubleunderline{78.75} & \doubleunderline{73.2} & \textbf{45.59} \\ 
SimMTM (p) & 95.12 & 98.52 & 96.39 & 77.68 & 59.07 & 43.73 & 95.46 & 50.26 & 80.37 & \textbf{79.79} & 72.95 & \doubleunderline{44.61} \\ 
Ci (s) & 95.12 & 98.49 & 98.58 & \doubleunderline{82.9} & \doubleunderline{59.98} & \textbf{46.51} & 95.32 & 50.26 & \underline{80.38} & 61.88 & \underline{73.12} & \underline{42.7} \\ 
Ci (p) & \doubleunderline{96.95} & 98.64 & 98.48 & 69.5 & 58.91 & 45.51 & 94.11 & \underline{50.26} & 80.38 & 60.83 & 72.31 & 42.14 \\ 
CiTrus (s) & 93.9 & 98.18 & 98.17 & \textbf{84.13} & \textbf{60.03} & \doubleunderline{46.5} & 95.28 & \doubleunderline{50.27} & \doubleunderline{80.38} & \underline{65.63} & 71.54 & 40.92 \\ 
CiTrus (p) & 95.12 & 98.38 & 98.45 & 80.93 & 59.53 & 45.7 & 90.28 & \textbf{50.27} & \textbf{80.38} & 65.21 & 72.31 & 41.16 \\ 
CiTrus (fp) & 54.27 & 76.11 & 72.11 & 30.84 & 53.11 & 36.88 & 56.54 & 49.99 & 80.23 & 38.54 & 60.53 & 32.32 \\ 
CiTrus (mp) & \underline{95.73} & 98.87 & 96.69 & \underline{81.79} & \underline{59.68} & \underline{45.89} & \textbf{95.55} & 50.25 & 80.37 & 62.29 & 71.96 & 40.85 \\ 

\hline
\end{tabular}
\caption{The performance of each model we use in this paper (trained by us) on the original test splits as defined in~\cite{zhang2022self}. The best performing models are made bold, second best are double-underlined, and third best are single-underlined.}
\label{tab:original-split}
\end{table*}
\FloatBarrier

\clearpage
\section{Frequency pre-training settings}
\label{app:frequency-pretraining}
Instead of predicting the masked signal in the original window, for the frequency pre-training strategy, the masked autoencoder model predicts the masked parts of the spectrogram instead.
To create the spectrogram from the pre-training data, we use Librosa's melspectrogram function~\cite{mcfee2015librosa}. Specifically, when pre-training on SleepEDF with $2$s windows, the spectrogram is generated with a hop\_length of $4$, $200$ ffts, $64$ mels, $fmin=0$, $fmax=20$, centering, and a window length of $60$. We add $40$ to the mel spectrogram and then divide them by $40$. For the $2$s pre-training, we only use the first $25$ spectrogram timesteps ($26$ are generated in total) because that corresponds to the number of patches after the convolutional encoder $200/8 = 25$. Lastly, the mel spectrograms are z-scored along the time dimension to push the model to focus on frequency content in the spectrograms that are masked that is not available in the other spectrograms. For the $30$s window pre-training setting, we use a hop\_length of $8$, $1024$ ffts, $128$ mels, and a window length of $60$. All other settings are the same. In this case we obtain $376$ timesteps, and since we have $3000/8=375$ patches after the convolutional encoder, we only use the first $375$.

\clearpage
\section{Result variance across seeds and folds}
\label{app:variance}
To verify that it is important to cross-validate transfer learning models for bio-signals, we evaluated how much the performance of each model fluctuates both across folds and across seeds. The variance across random seeds evaluates how models differ in their final solution with different starting and training stochasticity. The variance across test folds evaluates how the model performs with differences in the training and test distributions. Given that we are especially interested in low-data regimes, the variance in test folds can be very high. For example, we can get lucky and select a really good 1\% of the dataset as our training and validation set that represents the test set well. With only 1\% of the data, the chance that the training and validation set do not represent the test set well is much higher than it is when we use more data in our training and validation set. In Table~\ref{tab:variance} we see that variance across folds is markedly higher than it is across random seeds, sometimes leading to almost 5000\% more variance than seen across random seeds. The way we calculate the values is that we take the variance across test folds, average it across the three metrics, and divide it by the average variance across the random seeds and subtract $1$ from that result, and multiply it by $100$ to get a percentage. This gives us the percentage increase in variance.
\FloatBarrier
\begin{table*}[ht]
\centering
\begin{tabular}{l|ccc|ccc|ccc|ccc|}
\hline 
 Dataset & \multicolumn{3}{c}{EMG} & \multicolumn{3}{c}{ECG} & \multicolumn{3}{c}{PPG} & \multicolumn{3}{c}{HAR} \\ 
 \hline 
 Data & 20\% & 50\% & 80\% & 0.5\% & 1\% & 2\% & 10\% & 20\% & 50\% & 10\% & 20\% & 50\% \\ 
 \hline 
PatchTST (s) & \textbf{+53} & \textbf{+404} & \textbf{+136} & \textbf{+1488} & \textbf{+1496} & \textbf{+4319} & \textbf{+1713} & \textbf{+440} & \textbf{+197} & \textbf{+101} & \textbf{+384} & \textbf{+393} \\ 
PatchTST (p) & \textbf{+121} & \textbf{+40} & \textbf{+86} & \textbf{+998} & \textbf{+4608} & \textbf{+1905} & \textbf{+290} & \textbf{+974} & \textbf{+137} & \textbf{+119} & \textbf{+144} & \textbf{+380} \\ 
bioFAME (s) & \textbf{+71} & \textbf{+152} & \textbf{+112} & \textbf{+2311} & \textbf{+3478} & \textbf{+3745} & \textbf{+816} & \textbf{+497} & \textbf{+208} & \textbf{+475} & \textbf{+210} & \textbf{+187} \\ 
bioFAME (mp) & \textbf{+96} & \textbf{+154} & \textbf{+73} & \textbf{+2218} & \textbf{+2290} & \textbf{+3199} & \textbf{+421} & \textbf{+374} & \textbf{+391} & \textbf{+35} & \textbf{+20} & \textbf{+178} \\ 
NLPatchTST (s) & \textbf{+45} & \textbf{+28} & \textbf{+251} & \textbf{+1010} & \textbf{+1894} & \textbf{+2246} & \textbf{+420} & \textbf{+367} & \textbf{+86} & \textbf{+258} & \textbf{+498} & \textbf{+827} \\ 
NLPatchTST (p) & \textbf{+21} & \textbf{+53} & \textbf{+165} & \textbf{+1097} & \textbf{+1086} & \textbf{+1613} & \textbf{+231} & \textbf{+417} & \textbf{+148} & \textbf{+115} & \textbf{+131} & \textbf{+618} \\ 
NLPatchTST (mp) & \textbf{+124} & \textbf{+66} & \textbf{+229} & \textbf{+1785} & \textbf{+2252} & \textbf{+2087} & \textbf{+419} & \textbf{+319} & \textbf{+148} & \textbf{+117} & \textbf{+190} & \textbf{+706} \\ 
SimMTM (s) & \textbf{+88} & \textbf{+123} & \textbf{+100} & \textbf{+1082} & \textbf{+454} & \textbf{+1608} & \textbf{+850} & \textbf{+237} & \textbf{+479} & \textbf{+555} & \textbf{+732} & \textbf{+429} \\ 
SimMTM (p) & -56 & \textbf{+238} & \textbf{+66} & \textbf{+725} & \textbf{+827} & \textbf{+1903} & \textbf{+206} & \textbf{+494} & \textbf{+562} & \textbf{+506} & \textbf{+556} & \textbf{+439} \\ 
Ci (s) & \textbf{+1018} & \textbf{+703} & \textbf{+684} & \textbf{+2402} & \textbf{+1369} & \textbf{+2479} & \textbf{+488} & \textbf{+526} & \textbf{+298} & \textbf{+916} & \textbf{+1564} & \textbf{+1627} \\ 
Ci (p) & \textbf{+251} & \textbf{+191} & \textbf{+268} & \textbf{+2179} & \textbf{+3485} & \textbf{+2269} & \textbf{+625} & \textbf{+682} & \textbf{+564} & \textbf{+793} & \textbf{+1431} & \textbf{+1084} \\ 
CiTrus (s) & \textbf{+104} & \textbf{+559} & \textbf{+137} & \textbf{+2162} & \textbf{+1068} & \textbf{+1638} & \textbf{+113} & \textbf{+149} & \textbf{+212} & \textbf{+205} & \textbf{+342} & \textbf{+1088} \\ 
CiTrus (p) & -15 & \textbf{+100} & \textbf{+250} & \textbf{+1923} & \textbf{+1114} & \textbf{+1518} & \textbf{+346} & \textbf{+423} & \textbf{+294} & \textbf{+142} & \textbf{+737} & \textbf{+750} \\ 
CiTrus (fp) & \textbf{+89} & \textbf{+110} & \textbf{+424} & \textbf{+365} & \textbf{+320} & \textbf{+354} & \textbf{+18} & \textbf{+15} & \textbf{+45} & \textbf{+398} & \textbf{+192} & \textbf{+100} \\ 
CiTrus (mp) & \textbf{+129} & \textbf{+117} & \textbf{+174} & \textbf{+763} & \textbf{+3840} & \textbf{+1347} & \textbf{+255} & \textbf{+193} & \textbf{+309} & \textbf{+104} & \textbf{+882} & \textbf{+887} \\ 
\hline 
 Dataset & \multicolumn{3}{c}{FDB} & \multicolumn{3}{c}{Epilepsy} & \multicolumn{3}{c}{Gesture} & \multicolumn{3}{c}{SleepInference} \\ 
 \hline 
 Data & 0.5\% & 1\% & 2\% & 1\% & 5\% & 10\% & 10\% & 20\% & 50\% & 0.5\% & 1\% & 2\% \\ 
 \hline 
PatchTST (s) & \textbf{+71} & \textbf{+252} & \textbf{+41} & \textbf{+85} & \textbf{+102} & \textbf{+261} & \textbf{+339} & \textbf{+320} & \textbf{+379} &  &  &  \\ 
PatchTST (p) & \textbf{+42} & \textbf{+189} & \textbf{+44} & \textbf{+20} & \textbf{+89} & \textbf{+234} & \textbf{+143} & \textbf{+231} & \textbf{+549} &  &  &  \\ 
bioFAME (s) & \textbf{+238} & \textbf{+132} & \textbf{+181} & \textbf{+103} & \textbf{+171} & \textbf{+260} & \textbf{+137} & \textbf{+499} & \textbf{+618} &  &  &  \\ 
bioFAME (mp) & \textbf{+217} & \textbf{+213} & \textbf{+29} & \textbf{+37} & \textbf{+16} & \textbf{+154} & \textbf{+61} & \textbf{+34} & \textbf{+22} &  &  &  \\ 
NLPatchTST (s) & \textbf{+77} & \textbf{+83} & \textbf{+59} & -12 & -24 & \textbf{+13} & \textbf{+253} & \textbf{+439} & \textbf{+426} &  &  &  \\ 
NLPatchTST (p) & \textbf{+57} & \textbf{+54} & \textbf{+26} & \textbf{+26} & \textbf{+43} & \textbf{+180} & \textbf{+258} & \textbf{+219} & \textbf{+447} &  &  &  \\ 
NLPatchTST (mp) & \textbf{+109} & \textbf{+161} & \textbf{+112} & \textbf{+101} & \textbf{+99} & \textbf{+239} & \textbf{+199} & \textbf{+290} & \textbf{+220} &  &  &  \\ 
SimMTM (s) & \textbf{+49} & \textbf{+9} & \textbf{+68} & \textbf{+22} & \textbf{+146} & \textbf{+169} & \textbf{+902} & \textbf{+168} & \textbf{+460} &  &  &  \\ 
SimMTM (p) & \textbf{+120} & \textbf{+119} & \textbf{+175} & \textbf{+226} & \textbf{+192} & \textbf{+164} & \textbf{+350} & \textbf{+262} & \textbf{+219} &  &  &  \\ 
Ci (s) & \textbf{+337} & \textbf{+268} & \textbf{+190} & \textbf{+36} & \textbf{+215} & \textbf{+419} & \textbf{+214} & \textbf{+231} & \textbf{+344} &  &  &  \\ 
Ci (p) & \textbf{+420} & \textbf{+230} & \textbf{+31} & \textbf{+226} & \textbf{+188} & \textbf{+376} & \textbf{+54} & \textbf{+256} & \textbf{+475} &  &  &  \\ 
CiTrus (s) & \textbf{+469} & \textbf{+161} & \textbf{+70} & \textbf{+134} & \textbf{+50} & \textbf{+256} & \textbf{+99} & \textbf{+245} & \textbf{+255} &  &  &  \\ 
CiTrus (p) & \textbf{+105} & -22 & -25 & \textbf{+8} & \textbf{+67} & \textbf{+150} & \textbf{+148} & \textbf{+89} & \textbf{+248} &  &  &  \\ 
CiTrus (fp) & \textbf{+25} & \textbf{+232} & \textbf{+17} & \textbf{+7} & \textbf{+109} & \textbf{+92} & \textbf{+14} & \textbf{+71} & \textbf{+74} &  &  &  \\ 
CiTrus (mp) & \textbf{+112} & \textbf{+7} & \textbf{+37} & \textbf{+20} & \textbf{+8} & \textbf{+131} & \textbf{+129} & \textbf{+126} & \textbf{+112} &  &  &  \\ 

\hline
\end{tabular}
\caption{The percentage increase in variance across test folds compared to the variance across models for the same test fold. The percentage increase is averaged across all metrics. All percentages larger than $0$ are made bold.}
\label{tab:variance}
\end{table*}
\FloatBarrier

The following few tables provide the standard deviation in our results, both for every data regime (low, middle, and high), but also for our results on pre-training, reported in Table~\ref{tab:pretraining}, multimodal pre-training, reported in Table~\ref{tab:multimodal}, and our fine-tuning approach, reported in Table~\ref{tab:sliding-window}.
\clearpage
\subsection{Low-data regime results standard deviations}
\FloatBarrier
\begin{table*}[ht]
    \begin{tabular}{l|ccc|ccc|ccc|ccc|}
    \hline 
 Dataset & \multicolumn{3}{c}{EMG@20\%} & \multicolumn{3}{c}{ECG@0.5\%} & \multicolumn{3}{c}{PPG@10\%} & \multicolumn{3}{c}{HAR@10\%} \\ 
 \hline 
 Metric & ACC & ROC & PRC & ACC & ROC & PRC & ACC & ROC & PRC & ACC & ROC & PRC \\ 
 \hline 
PatchTST (s) & 8.59 & 11.1 & 11.15 & 11.71 & 1.94 & 17.44 & 7.96 & 6.31 & 7.12 & 4.58 & 2.89 & 4.9 \\ 
PatchTST (p) & 11.2 & 11.94 & 13.38 & 13.27 & 2.22 & 17.45 & 7.46 & 7.71 & 9.12 & 5.02 & 2.42 & 3.95 \\ 
bioFAME (s) & 9.08 & 13.04 & 10.56 & 7.92 & 1.6 & 18.44 & 5.98 & 4.24 & 5.91 & 5.28 & 2.2 & 3.47 \\ 
bioFAME (mp) & 8.35 & 6.91 & 10.62 & 11.47 & 2.89 & 18.15 & 7.11 & 5.16 & 7.04 & 5.86 & 3.18 & 5.34 \\ 
NLPatchTST (s) & 13.32 & 8.99 & 10.99 & 14.6 & 1.22 & 17.61 & 8.51 & 7.42 & 7.46 & 5.17 & 2.69 & 4.97 \\ 
NLPatchTST (p) & 11.7 & 7.33 & 10.21 & 14.66 & 2.02 & 17.55 & 6.22 & 6.19 & 7.04 & 6.1 & 2.67 & 5.37 \\ 
NLPatchTST (mp) & 10.81 & 6.54 & 9.66 & 14.42 & 2.0 & 17.4 & 6.89 & 5.9 & 6.8 & 5.37 & 2.51 & 4.6 \\ 
SimMTM (s) & 21.42 & 10.76 & 15.44 & 17.71 & 2.19 & 17.08 & 10.77 & 9.55 & 12.36 & 5.75 & 2.64 & 5.15 \\ 
SimMTM (p) & 6.43 & 17.95 & 14.01 & 15.16 & 1.98 & 17.39 & 10.78 & 10.59 & 14.06 & 4.19 & 2.39 & 4.78 \\ 
Ci (s) & 4.47 & 2.25 & 5.25 & 17.11 & 3.67 & 17.43 & 8.27 & 7.79 & 7.47 & 3.34 & 2.64 & 4.78 \\ 
Ci (p) & 5.29 & 2.19 & 4.99 & 17.89 & 3.13 & 17.23 & 7.54 & 7.7 & 7.69 & 6.46 & 2.8 & 5.62 \\ 
CiTrus (s) & 4.85 & 1.13 & 1.18 & 15.4 & 3.7 & 17.37 & 7.76 & 8.57 & 9.45 & 4.63 & 2.47 & 4.96 \\ 
CiTrus (p) & 8.04 & 4.41 & 7.47 & 18.86 & 3.1 & 17.42 & 6.72 & 8.21 & 8.67 & 4.7 & 2.67 & 5.31 \\ 
CiTrus (fp) & 3.93 & 3.38 & 3.73 & 11.84 & 4.26 & 17.51 & 8.3 & 9.24 & 11.74 & 5.02 & 2.64 & 4.83 \\ 
CiTrus (mp) & 6.17 & 1.4 & 4.67 & 16.94 & 3.43 & 17.74 & 5.85 & 6.24 & 6.32 & 5.43 & 2.82 & 5.63 \\ 
\hline 
 Dataset & \multicolumn{3}{c}{FDB@0.5\%} & \multicolumn{3}{c}{Epilepsy@1\%} & \multicolumn{3}{c}{Gesture@10\%} & \multicolumn{3}{c}{SleepInference@0.5\%} \\ 
 \hline 
 Metric & ACC & ROC & PRC & ACC & ROC & PRC & ACC & ROC & PRC & ACC & ROC & PRC \\ 
 \hline 
PatchTST (s) & 3.1 & 1.64 & 1.19 & 1.42 & 4.04 & 2.3 & 7.68 & 3.64 & 7.24 & 0.92 & 0.35 & 0.51 \\ 
PatchTST (p) & 2.92 & 1.84 & 1.35 & 1.06 & 2.3 & 1.32 & 8.11 & 3.34 & 7.08 & 1.28 & 0.75 & 0.96 \\ 
bioFAME (s) & 3.94 & 1.99 & 1.3 & 1.8 & 1.76 & 0.82 & 7.33 & 4.81 & 5.73 & 0.39 & 0.58 & 0.65 \\ 
bioFAME (mp) & 5.05 & 2.83 & 2.24 & 1.71 & 1.0 & 0.4 & 7.79 & 6.03 & 7.55 & 2.07 & 0.46 & 0.81 \\ 
NLPatchTST (s) & 5.84 & 2.37 & 1.59 & 4.5 & 9.69 & 4.85 & 9.28 & 4.45 & 7.9 & 2.22 & 1.09 & 1.18 \\ 
NLPatchTST (p) & 2.37 & 1.76 & 1.26 & 1.16 & 3.31 & 2.1 & 7.88 & 4.31 & 7.94 & 1.03 & 0.53 & 1.07 \\ 
NLPatchTST (mp) & 2.59 & 1.98 & 1.57 & 1.5 & 3.72 & 2.34 & 6.49 & 3.72 & 7.05 & 2.29 & 0.81 & 1.21 \\ 
SimMTM (s) & 7.45 & 2.68 & 2.56 & 3.19 & 0.24 & 1.36 & 9.09 & 4.1 & 8.09 & 0.0 & 0.0 & 0.0 \\ 
SimMTM (p) & 12.16 & 5.89 & 4.66 & 0.92 & 0.32 & 1.34 & 7.68 & 3.23 & 7.1 & 0.0 & 0.0 & 0.0 \\ 
Ci (s) & 8.19 & 2.97 & 3.33 & 2.07 & 0.73 & 0.32 & 7.53 & 6.02 & 8.27 & 1.52 & 0.43 & 0.68 \\ 
Ci (p) & 6.7 & 3.12 & 3.44 & 0.92 & 0.69 & 0.28 & 7.19 & 4.22 & 6.73 & 1.1 & 0.53 & 0.64 \\ 
CiTrus (s) & 7.89 & 3.45 & 3.73 & 1.84 & 0.95 & 0.51 & 7.32 & 4.43 & 6.83 & 2.25 & 0.62 & 0.76 \\ 
CiTrus (p) & 4.82 & 2.08 & 2.85 & 1.34 & 4.15 & 2.53 & 8.47 & 4.51 & 7.79 & 0.78 & 0.26 & 0.82 \\ 
CiTrus (fp) & 5.71 & 2.13 & 2.47 & 1.8 & 1.18 & 0.59 & 7.18 & 4.81 & 7.51 & 5.31 & 2.18 & 2.3 \\ 
CiTrus (mp) & 3.77 & 2.41 & 2.82 & 1.0 & 4.38 & 2.82 & 6.7 & 3.85 & 6.71 & 3.3 & 1.14 & 1.62 \\ 
\hline
\end{tabular}
\caption{The standard-deviations across both seeds and folds for the low-data regime.}
\label{tab:low-data-variance}
\end{table*}
\FloatBarrier

\clearpage
\subsection{Middle-data regime results standard deviations}
\FloatBarrier
\begin{table*}[ht]
    \begin{tabular}{l|ccc|ccc|ccc|ccc|}
    \hline 
 Dataset & \multicolumn{3}{c}{EMG@50\%} & \multicolumn{3}{c}{ECG@1\%} & \multicolumn{3}{c}{PPG@20\%} & \multicolumn{3}{c}{HAR@20\%} \\ 
 \hline 
 Metric & ACC & ROC & PRC & ACC & ROC & PRC & ACC & ROC & PRC & ACC & ROC & PRC \\ 
 \hline 
PatchTST (s) & 12.55 & 10.73 & 14.25 & 15.05 & 2.41 & 16.76 & 8.16 & 5.08 & 6.27 & 3.86 & 2.66 & 4.8 \\ 
PatchTST (p) & 12.75 & 13.02 & 14.71 & 14.13 & 2.62 & 17.38 & 8.9 & 6.47 & 7.21 & 5.59 & 2.7 & 5.01 \\ 
bioFAME (s) & 12.44 & 14.84 & 15.79 & 9.84 & 2.48 & 18.09 & 8.12 & 3.64 & 4.03 & 4.5 & 2.58 & 3.88 \\ 
bioFAME (mp) & 7.14 & 3.52 & 7.57 & 12.26 & 2.96 & 18.31 & 7.94 & 4.48 & 7.11 & 6.47 & 3.02 & 5.59 \\ 
NLPatchTST (s) & 7.97 & 4.23 & 6.94 & 15.43 & 2.15 & 16.99 & 10.96 & 6.36 & 8.68 & 4.84 & 2.86 & 5.85 \\ 
NLPatchTST (p) & 5.76 & 2.88 & 5.18 & 16.64 & 2.17 & 17.09 & 7.47 & 6.34 & 7.27 & 5.49 & 2.74 & 5.42 \\ 
NLPatchTST (mp) & 10.13 & 4.3 & 7.02 & 15.92 & 2.38 & 17.46 & 8.74 & 5.16 & 6.25 & 6.46 & 2.55 & 5.11 \\ 
SimMTM (s) & 6.51 & 3.09 & 7.08 & 13.0 & 3.31 & 17.14 & 10.41 & 8.04 & 10.97 & 5.35 & 3.04 & 5.8 \\ 
SimMTM (p) & 8.78 & 6.8 & 9.31 & 15.76 & 3.22 & 16.96 & 9.74 & 8.29 & 10.78 & 4.12 & 3.1 & 5.86 \\ 
Ci (s) & 3.26 & 2.34 & 5.42 & 15.92 & 3.49 & 16.88 & 7.6 & 6.39 & 7.56 & 3.11 & 2.69 & 5.03 \\ 
Ci (p) & 5.03 & 1.64 & 3.38 & 18.42 & 3.5 & 16.74 & 8.24 & 7.22 & 7.99 & 6.26 & 2.88 & 5.61 \\ 
CiTrus (s) & 3.27 & 1.28 & 0.86 & 16.79 & 3.85 & 17.2 & 7.36 & 6.14 & 6.28 & 4.53 & 2.68 & 4.62 \\ 
CiTrus (p) & 4.84 & 1.56 & 4.16 & 20.0 & 3.59 & 17.01 & 9.04 & 8.25 & 8.66 & 4.2 & 2.64 & 5.33 \\ 
CiTrus (fp) & 4.09 & 1.49 & 3.47 & 12.01 & 4.3 & 17.53 & 7.04 & 7.4 & 9.59 & 6.5 & 2.53 & 4.46 \\ 
CiTrus (mp) & 5.61 & 3.28 & 5.7 & 17.04 & 3.89 & 17.12 & 8.89 & 7.09 & 7.88 & 4.9 & 2.85 & 5.57 \\ 
\hline 
 Dataset & \multicolumn{3}{c}{FDB@1\%} & \multicolumn{3}{c}{Epilepsy@5\%} & \multicolumn{3}{c}{Gesture@20\%} & \multicolumn{3}{c}{SleepInference@1\%} \\ 
 \hline 
 Metric & ACC & ROC & PRC & ACC & ROC & PRC & ACC & ROC & PRC & ACC & ROC & PRC \\ 
 \hline 
PatchTST (s) & 5.96 & 2.46 & 2.07 & 0.7 & 0.89 & 0.4 & 7.19 & 3.63 & 7.79 & 1.94 & 0.37 & 0.25 \\ 
PatchTST (p) & 3.4 & 1.64 & 1.36 & 0.85 & 0.96 & 0.55 & 6.83 & 2.62 & 5.51 & 1.74 & 0.25 & 0.43 \\ 
bioFAME (s) & 4.59 & 2.7 & 2.01 & 0.79 & 0.71 & 0.25 & 8.53 & 4.81 & 8.06 & 0.84 & 0.12 & 0.13 \\ 
bioFAME (mp) & 4.04 & 2.51 & 2.2 & 0.86 & 0.68 & 0.26 & 8.04 & 4.13 & 8.58 & 2.22 & 0.29 & 0.61 \\ 
NLPatchTST (s) & 4.86 & 2.44 & 2.4 & 5.52 & 14.84 & 7.48 & 7.54 & 3.91 & 8.05 & 2.58 & 0.45 & 0.35 \\ 
NLPatchTST (p) & 4.43 & 2.2 & 2.01 & 1.07 & 2.18 & 1.29 & 7.13 & 2.69 & 6.61 & 0.26 & 0.17 & 0.43 \\ 
NLPatchTST (mp) & 4.15 & 2.6 & 2.07 & 0.86 & 1.73 & 0.96 & 6.8 & 2.93 & 5.91 & 1.42 & 0.23 & 0.6 \\ 
SimMTM (s) & 12.55 & 5.6 & 5.26 & 0.68 & 0.18 & 1.38 & 5.56 & 1.99 & 4.31 & 0.0 & 0.0 & 0.0 \\ 
SimMTM (p) & 1.7 & 1.28 & 1.18 & 0.84 & 0.25 & 1.19 & 4.73 & 2.06 & 4.08 & 0.0 & 0.0 & 0.0 \\ 
Ci (s) & 2.96 & 1.02 & 1.23 & 0.82 & 0.5 & 0.24 & 7.11 & 3.06 & 7.41 & 1.14 & 0.16 & 0.43 \\ 
Ci (p) & 3.01 & 1.21 & 2.14 & 0.65 & 0.42 & 0.19 & 7.33 & 3.32 & 7.2 & 0.84 & 0.23 & 0.3 \\ 
CiTrus (s) & 4.72 & 1.1 & 2.19 & 0.88 & 0.6 & 0.3 & 6.78 & 2.39 & 5.69 & 0.56 & 0.16 & 0.43 \\ 
CiTrus (p) & 3.8 & 1.92 & 2.79 & 0.77 & 1.69 & 1.14 & 7.19 & 2.85 & 6.87 & 1.34 & 0.15 & 0.3 \\ 
CiTrus (fp) & 1.29 & 1.25 & 1.61 & 0.68 & 0.44 & 0.23 & 8.91 & 3.63 & 8.12 & 8.14 & 4.16 & 3.31 \\ 
CiTrus (mp) & 3.56 & 1.54 & 2.31 & 0.83 & 1.5 & 1.1 & 6.13 & 2.81 & 5.63 & 1.59 & 0.24 & 0.6 \\ 
\hline
\end{tabular}
\caption{The standard-deviations across both seeds and folds for the middle-data regime.}
\label{tab:mid-data-variance}
\end{table*}
\FloatBarrier

\clearpage
\subsection{High-data regime results standard deviations}
\FloatBarrier
\begin{table*}[ht]
    \begin{tabular}{l|ccc|ccc|ccc|ccc|}
    \hline 
 Dataset & \multicolumn{3}{c}{EMG@80\%} & \multicolumn{3}{c}{ECG@2\%} & \multicolumn{3}{c}{PPG@50\%} & \multicolumn{3}{c}{HAR@50\%} \\ 
 \hline 
 Metric & ACC & ROC & PRC & ACC & ROC & PRC & ACC & ROC & PRC & ACC & ROC & PRC \\ 
 \hline 
PatchTST (s) & 6.95 & 2.87 & 5.52 & 14.31 & 2.6 & 17.13 & 7.41 & 4.52 & 5.49 & 3.84 & 2.58 & 4.91 \\ 
PatchTST (p) & 8.47 & 6.85 & 9.51 & 15.6 & 2.39 & 17.18 & 6.52 & 5.21 & 6.87 & 4.07 & 2.66 & 5.04 \\ 
bioFAME (s) & 8.81 & 5.66 & 9.0 & 7.79 & 2.2 & 18.31 & 5.76 & 3.45 & 4.46 & 5.22 & 2.56 & 4.42 \\ 
bioFAME (mp) & 6.98 & 2.79 & 5.69 & 8.34 & 2.96 & 18.41 & 6.2 & 7.37 & 9.98 & 4.48 & 2.91 & 5.27 \\ 
NLPatchTST (s) & 5.99 & 1.77 & 4.32 & 17.17 & 2.76 & 17.69 & 7.98 & 4.69 & 7.25 & 3.28 & 2.53 & 4.93 \\ 
NLPatchTST (p) & 5.18 & 2.92 & 6.23 & 15.65 & 2.08 & 17.54 & 5.87 & 4.38 & 6.12 & 3.29 & 2.74 & 5.49 \\ 
NLPatchTST (mp) & 6.13 & 3.21 & 7.26 & 14.15 & 2.48 & 17.8 & 4.31 & 4.14 & 5.19 & 3.73 & 2.63 & 5.03 \\ 
SimMTM (s) & 5.93 & 1.45 & 3.51 & 17.75 & 2.79 & 16.67 & 10.27 & 7.74 & 11.1 & 4.61 & 2.53 & 5.13 \\ 
SimMTM (p) & 6.91 & 2.27 & 3.9 & 16.9 & 2.17 & 16.41 & 8.94 & 7.63 & 10.79 & 4.08 & 2.66 & 5.29 \\ 
Ci (s) & 3.05 & 1.46 & 3.45 & 20.55 & 3.7 & 17.0 & 6.72 & 5.75 & 7.89 & 2.41 & 2.51 & 4.54 \\ 
Ci (p) & 3.9 & 1.61 & 2.93 & 14.74 & 3.77 & 17.77 & 5.75 & 6.05 & 7.66 & 3.73 & 2.58 & 4.74 \\ 
CiTrus (s) & 3.67 & 1.02 & 0.7 & 14.44 & 3.87 & 17.24 & 7.48 & 6.66 & 10.59 & 2.78 & 2.6 & 5.01 \\ 
CiTrus (p) & 3.51 & 1.88 & 2.31 & 16.83 & 3.95 & 17.35 & 5.75 & 7.17 & 8.31 & 3.05 & 2.59 & 5.08 \\ 
CiTrus (fp) & 3.32 & 0.99 & 0.84 & 10.68 & 4.33 & 17.59 & 6.31 & 6.42 & 9.48 & 4.28 & 2.59 & 4.88 \\ 
CiTrus (mp) & 3.94 & 1.29 & 3.01 & 14.7 & 3.89 & 17.65 & 5.65 & 5.76 & 8.13 & 3.3 & 2.69 & 5.27 \\ 
\hline 
 Dataset & \multicolumn{3}{c}{FDB@2\%} & \multicolumn{3}{c}{Epilepsy@10\%} & \multicolumn{3}{c}{Gesture@50\%} & \multicolumn{3}{c}{SleepInference@2\%} \\ 
 \hline 
 Metric & ACC & ROC & PRC & ACC & ROC & PRC & ACC & ROC & PRC & ACC & ROC & PRC \\ 
 \hline 
PatchTST (s) & 4.68 & 2.27 & 2.42 & 0.77 & 0.79 & 0.31 & 6.92 & 3.04 & 7.71 & 1.01 & 0.19 & 0.26 \\ 
PatchTST (p) & 3.24 & 1.92 & 1.76 & 0.77 & 0.94 & 0.5 & 7.44 & 2.66 & 6.82 & 1.76 & 0.53 & 0.59 \\ 
bioFAME (s) & 4.32 & 2.4 & 2.44 & 0.88 & 0.66 & 0.23 & 8.37 & 3.56 & 8.52 & 0.94 & 0.25 & 0.31 \\ 
bioFAME (mp) & 2.26 & 1.61 & 2.13 & 0.69 & 0.5 & 0.22 & 7.42 & 3.31 & 7.57 & 2.89 & 0.72 & 1.16 \\ 
NLPatchTST (s) & 3.48 & 2.28 & 2.51 & 4.51 & 5.43 & 3.22 & 6.43 & 3.51 & 7.73 & 2.73 & 0.98 & 0.92 \\ 
NLPatchTST (p) & 4.02 & 2.31 & 2.33 & 0.74 & 0.81 & 0.38 & 7.84 & 2.86 & 7.81 & 0.88 & 0.09 & 0.17 \\ 
NLPatchTST (mp) & 2.49 & 1.87 & 1.87 & 0.8 & 0.84 & 0.37 & 7.8 & 3.18 & 7.51 & 1.32 & 0.4 & 0.73 \\ 
SimMTM (s) & 2.67 & 1.16 & 1.18 & 0.39 & 0.29 & 1.09 & 5.04 & 2.11 & 5.21 & 0.0 & 0.0 & 0.0 \\ 
SimMTM (p) & 1.45 & 1.14 & 1.18 & 0.59 & 0.24 & 0.8 & 5.07 & 1.75 & 3.84 & 0.0 & 0.0 & 0.0 \\ 
Ci (s) & 2.52 & 1.13 & 1.14 & 0.88 & 0.32 & 0.1 & 5.72 & 2.56 & 5.88 & 2.75 & 0.44 & 0.41 \\ 
Ci (p) & 2.18 & 1.14 & 1.11 & 0.73 & 0.31 & 0.13 & 7.05 & 2.79 & 5.7 & 0.92 & 0.12 & 0.12 \\ 
CiTrus (s) & 4.27 & 1.2 & 1.91 & 0.78 & 0.35 & 0.12 & 5.84 & 2.47 & 5.4 & 1.0 & 0.49 & 0.8 \\ 
CiTrus (p) & 3.85 & 1.51 & 2.17 & 0.84 & 0.52 & 0.3 & 7.05 & 2.85 & 7.53 & 1.52 & 0.35 & 0.61 \\ 
CiTrus (fp) & 1.69 & 1.16 & 1.66 & 0.61 & 0.36 & 0.14 & 10.78 & 3.07 & 8.8 & 5.61 & 1.83 & 2.25 \\ 
CiTrus (mp) & 3.14 & 1.42 & 1.85 & 0.78 & 0.91 & 0.69 & 5.81 & 2.81 & 5.82 & 1.22 & 0.73 & 1.22 \\ 
\hline
\end{tabular}
\caption{The standard-deviations across both seeds and folds for the high-data regime.}
\label{tab:high-data-variance}
\end{table*}
\FloatBarrier

\clearpage
\subsection{Pre-training results standard deviations}
\FloatBarrier
\begin{table*}[ht]
\centering
\begin{tabular}{l|ccc|ccc|ccc|ccc|}
\hline 
 Dataset & \multicolumn{3}{c}{EMG} & \multicolumn{3}{c}{ECG} & \multicolumn{3}{c}{PPG} & \multicolumn{3}{c}{HAR} \\ 
 \hline 
 Data & 20\% & 50\% & 80\% & 0.5\% & 1\% & 2\% & 10\% & 20\% & 50\% & 10\% & 20\% & 50\% \\ 
 \hline 
PatchTST & 13.2 & 13.0 & 12.8 & 4.4 & 4.2 & 3.8 & 6.0 & 4.8 & 6.4 & 4.4 & 3.7 & 2.8 \\ 
bioFAME & 12.8 & 10.3 & 14.1 & 4.1 & 3.6 & 3.0 & 4.7 & 5.2 & 8.5 & 4.4 & 4.0 & 4.5 \\ 
NLPatchTST (p) & 15.1 & 15.3 & 15.3 & 4.4 & 4.8 & 4.6 & 6.0 & 5.9 & 6.9 & 3.8 & 4.2 & 2.2 \\ 
NLPatchTST (mp) & 13.2 & 11.7 & 14.4 & 4.8 & 4.4 & 3.9 & 5.6 & 6.6 & 7.1 & 3.7 & 4.5 & 2.0 \\ 
Ci & 14.6 & 10.1 & 5.4 & 4.3 & 4.5 & 5.3 & 5.2 & 5.1 & 5.3 & 3.5 & 3.1 & 1.5 \\ 
CiTrus (p) & 12.8 & 10.3 & 5.9 & 5.2 & 4.8 & 4.1 & 8.8 & 7.4 & 9.1 & 4.6 & 2.7 & 1.7 \\ 
CiTrus (fp) & 15.1 & 7.7 & 5.5 & 5.3 & 5.8 & 4.9 & 11.2 & 8.7 & 8.9 & 4.0 & 4.2 & 2.2 \\ 
CiTrus (mp) & 14.6 & 9.6 & 6.6 & 5.2 & 3.8 & 4.1 & 8.5 & 7.0 & 8.4 & 4.7 & 2.5 & 1.9 \\ 
\hline 
 Dataset & \multicolumn{3}{c}{FDB} & \multicolumn{3}{c}{Epilepsy} & \multicolumn{3}{c}{Gesture} & \multicolumn{3}{c}{SleepInference} \\ 
 \hline 
 Data & 0.5\% & 1\% & 2\% & 1\% & 5\% & 10\% & 10\% & 20\% & 50\% & 0.5\% & 1\% & 2\% \\ 
 \hline 
PatchTST & 2.0 & 1.8 & 1.4 & 2.5 & 0.9 & 0.8 & 6.2 & 5.1 & 4.4 & 1.0 & 0.6 & 0.9 \\ 
bioFAME & 1.7 & 1.6 & 0.9 & 1.5 & 0.6 & 0.4 & 7.0 & 6.2 & 6.3 & 1.4 & 1.0 & 1.5 \\ 
NLPatchTST (p) & 1.9 & 2.0 & 1.3 & 6.6 & 9.3 & 4.4 & 6.6 & 5.4 & 4.6 & 1.7 & 1.1 & 1.4 \\ 
NLPatchTST (mp) & 2.3 & 2.0 & 1.3 & 6.6 & 9.6 & 4.3 & 6.5 & 5.1 & 4.3 & 2.2 & 1.1 & 1.1 \\ 
Ci & 1.9 & 1.6 & 1.4 & 1.0 & 0.4 & 0.3 & 7.4 & 4.5 & 3.6 & 1.5 & 1.0 & 1.2 \\ 
CiTrus (p) & 1.5 & 1.5 & 1.2 & 3.1 & 1.3 & 0.6 & 7.7 & 6.1 & 5.2 & 1.7 & 0.7 & 1.3 \\ 
CiTrus (fp) & 2.0 & 1.6 & 1.8 & 1.5 & 0.6 & 0.4 & 8.0 & 6.6 & 6.9 & 3.7 & 5.3 & 3.8 \\ 
CiTrus (mp) & 2.4 & 1.7 & 1.3 & 3.1 & 1.3 & 0.8 & 6.6 & 4.6 & 4.1 & 2.5 & 0.8 & 1.5 \\ 

\hline
\end{tabular}
\caption{The standard-deviations across both seeds and folds in the average performance improvement results when pre-training in Table~\ref{tab:pretraining}.}
\label{tab:pretrain-variance}
\end{table*}
\FloatBarrier

\subsection{Multi-modal pre-training results standard deviations}
\FloatBarrier
\begin{table*}[ht]
\centering
\begin{tabular}{l|ccc|ccc|ccc|ccc|}
\hline 
 Dataset & \multicolumn{3}{c}{EMG} & \multicolumn{3}{c}{ECG} & \multicolumn{3}{c}{PPG} & \multicolumn{3}{c}{HAR} \\ 
 \hline 
 Data & 20\% & 50\% & 80\% & 0.5\% & 1\% & 2\% & 10\% & 20\% & 50\% & 10\% & 20\% & 50\% \\ 
 \hline 
NLPatchTST & 32.0 & 22.8 & 20.5 & 7.6 & 12.3 & 7.4 & 9.6 & 9.2 & 9.4 & 6.0 & 5.6 & 2.3 \\ 
CiTrus & 17.1 & 17.1 & 8.8 & 25.4 & 15.5 & 8.4 & 9.4 & 11.7 & 8.6 & 4.9 & 2.9 & 2.2 \\ 
\hline 
 Dataset & \multicolumn{3}{c}{FDB} & \multicolumn{3}{c}{Epilepsy} & \multicolumn{3}{c}{Gesture} & \multicolumn{3}{c}{SleepInference} \\ 
 \hline 
 Data & 0.5\% & 1\% & 2\% & 1\% & 5\% & 10\% & 10\% & 20\% & 50\% & 0.5\% & 1\% & 2\% \\ 
 \hline 
NLPatchTST & 4.2 & 2.9 & 1.7 & 3.1 & 1.6 & 0.6 & 11.3 & 6.4 & 6.6 & 1.6 & 1.5 & 1.2 \\ 
CiTrus & 4.0 & 2.8 & 2.5 & 3.5 & 1.6 & 0.8 & 17.6 & 11.7 & 7.8 & 4.0 & 1.6 & 2.7 \\ 

\hline
\end{tabular}
\caption{The standard-deviations across both seeds and folds in the average performance improvement results with multimodal pre-training in Table~\ref{tab:multimodal}.}
\label{tab:multimodal-variance}
\end{table*}
\FloatBarrier

\clearpage
\subsection{Fine-tuning approach results standard deviations}
\FloatBarrier
\begin{table*}[ht]
\centering
\begin{tabular}{l|ccc|ccc|ccc|ccc|}
\hline 
 Dataset & \multicolumn{3}{c}{ECG} & \multicolumn{3}{c}{PPG} & \multicolumn{3}{c}{HAR} & \multicolumn{3}{c}{Gesture} \\ 
 \hline 
 Data & 0.5\% & 1\% & 2\% & 10\% & 20\% & 50\% & 10\% & 20\% & 50\% & 10\% & 20\% & 50\% \\ 
 \hline 
PatchTST (s) & 15.5 & 11.1 & 8.2 & 28.7 & 31.1 & 26.6 & 7.0 & 6.2 & 4.2 & 24.2 & 13.4 & 9.4 \\ 
PatchTST (p) & 13.9 & 18.1 & 9.6 & 28.1 & 30.3 & 27.5 & 8.9 & 7.8 & 5.4 & 10.5 & 8.4 & 7.9 \\ 
bioFAME (s) & 13.1 & 10.1 & 8.4 & 37.5 & 30.3 & 22.2 & 7.3 & 6.1 & 6.1 & 17.2 & 12.0 & 10.7 \\ 
bioFAME (mp) & 20.6 & 13.2 & 12.2 & 31.8 & 24.2 & 35.4 & 7.3 & 7.0 & 5.6 & 15.6 & 9.8 & 10.1 \\ 
SimMTM (s) & 17.1 & 17.8 & 14.3 & 28.7 & 24.0 & 39.2 & 3.3 & 2.5 & 2.4 & 6.4 & 5.8 & 4.4 \\ 
SimMTM (p) & 12.6 & 9.7 & 12.3 & 28.7 & 24.8 & 25.4 & 1.7 & 1.5 & 1.9 & 13.0 & 3.8 & 3.3 \\ 
NLPatchTST (s) & 10.6 & 9.5 & 8.7 & 22.2 & 30.5 & 27.5 & 6.8 & 6.6 & 4.0 & 18.4 & 13.4 & 7.0 \\ 
NLPatchTST (p) & 13.6 & 16.0 & 12.9 & 26.6 & 39.1 & 27.6 & 7.6 & 7.1 & 5.6 & 8.6 & 6.8 & 7.7 \\ 
NLPatchTST (mp) & 12.5 & 21.9 & 8.3 & 25.5 & 33.3 & 28.3 & 7.6 & 7.6 & 5.9 & 9.8 & 8.5 & 8.1 \\ 
Ci (s) & 11.9 & 12.8 & 13.6 & 15.9 & 27.5 & 20.8 & 6.9 & 4.9 & 3.6 & 30.9 & 8.8 & 7.7 \\ 
Ci (p) & 11.9 & 20.4 & 17.8 & 20.0 & 32.4 & 24.0 & 6.2 & 5.5 & 4.5 & 12.8 & 7.5 & 7.1 \\ 
CiTrus (s) & 9.5 & 18.5 & 8.1 & 23.5 & 25.7 & 19.9 & 8.6 & 4.9 & 3.8 & 33.1 & 10.2 & 6.6 \\ 
CiTrus (p) & 33.8 & 25.5 & 25.7 & 29.6 & 29.4 & 31.8 & 6.8 & 4.9 & 4.6 & 15.2 & 8.0 & 7.6 \\ 
CiTrus (fp) & 6.1 & 18.8 & 8.9 & 24.7 & 24.4 & 23.7 & 4.8 & 6.9 & 4.9 & 13.3 & 7.8 & 6.7 \\ 
CiTrus (mp) & 44.1 & 22.3 & 20.3 & 27.5 & 31.6 & 29.6 & 7.7 & 5.4 & 4.0 & 10.0 & 8.6 & 6.8 \\ 

\hline
\end{tabular}
\caption{The standard-deviations across both seeds and folds in the average performance improvement results when using our fine-tuning approach in Table~\ref{tab:sliding-window}.}
\label{tab:sliding-window-variance}
\end{table*}
\FloatBarrier

\clearpage
\section{Average model results}
\label{app:avg-results}
The following table, Table~\ref{tab:avg-performance}, shows the average performance of each model across the datasets. The results are shown for each metric, and for each data regime. Generally, the best-performing models are variants of our proposed model architectures. Especially CiTrus (fp), and CiTrus (mp) perform really well for the low-data regime. CiTrus (fp) clearly performs the best for the high data regime. For the middle data regie, the best model is Ci (s).

\FloatBarrier
\begin{table*}[ht]
\centering
\begin{tabular}{l|ccc|ccc|ccc|}
\hline 
 Data regime & \multicolumn{3}{c}{Low} & \multicolumn{3}{c}{Middle} & \multicolumn{3}{c}{High} \\ 
 \hline 
 Metric & ACC & ROC & PRC & ACC & ROC & PRC & ACC & ROC & PRC \\ 
 \hline 
PatchTST (s) & 60.79 & 71.31 & 56.97 & 67.48 & 76.89 & 63.82 & 73.88 & 79.65 & 68.49 \\ 
PatchTST (p) & 63.63 & 73.25 & 59.02 & 67.39 & 75.83 & 62.46 & 73.1 & 78.76 & 67.05 \\ 
bioFAME (s) & 58.58 & 70.52 & 54.84 & 63.0 & 74.54 & 59.52 & 68.05 & 77.59 & 63.1 \\ 
bioFAME (mp) & 64.04 & 75.54 & 60.89 & 67.85 & 77.5 & 63.73 & 73.33 & 79.46 & 67.23 \\ 
NLPatchTST (s) & 60.86 & 72.01 & 58.5 & 68.77 & 75.89 & 64.01 & 74.71 & 79.53 & 68.8 \\ 
NLPatchTST (p) & 66.18 & 75.78 & 61.94 & 70.82 & 78.13 & 65.94 & 76.04 & 80.12 & 69.35 \\ 
NLPatchTST (mp) & 65.99 & 75.71 & 61.7 & 70.66 & 77.99 & 65.47 & 75.93 & 80.13 & 69.21 \\ 
Ci (s) & 71.9 & 78.45 & \doubleunderline{66.39} & \textbf{78.37} & \textbf{81.19} & \textbf{70.33} & 80.21 & \doubleunderline{82.4} & \doubleunderline{72.42} \\ 
Ci (p) & 70.94 & \textbf{78.7} & 65.74 & 76.12 & \doubleunderline{80.72} & 69.19 & \doubleunderline{80.78} & \underline{82.04} & 71.96 \\ 
CiTrus (s) & 71.52 & \underline{78.53} & 66.08 & \doubleunderline{77.15} & \underline{80.65} & \doubleunderline{69.37} & \underline{80.68} & 82.0 & \underline{72.1} \\ 
CiTrus (p) & \underline{71.93} & 78.28 & 65.78 & 75.89 & 80.18 & 68.81 & 80.38 & 81.81 & 71.7 \\ 
CiTrus (fp) & \textbf{73.01} & 78.1 & \underline{66.18} & \underline{76.8} & 80.36 & 69.08 & \textbf{82.6} & \textbf{82.52} & \textbf{72.91} \\ 
CiTrus (mp) & \doubleunderline{72.71} & \doubleunderline{78.69} & \textbf{66.42} & 76.67 & 80.19 & \underline{69.22} & 80.28 & 81.4 & 71.15 \\ 

\hline
\end{tabular}
\caption{A comparison of the different model architectures on average across the datasets for each data regime. The best result for each metric is shown in bold, the second best result is double-underlined, and the third best result is single-underlined.}
\label{tab:avg-performance}
\end{table*}
\FloatBarrier

\clearpage

\section{Experimental statistics}
\label{app:experimental-statistics}
To statistically verify the experiments in the main paper, we perform a one-sided Wilcoxon statistical test. For Table~\ref{tab:architecture-significance} we compare each model's performance to our CiTrus (fp) model and compute whether our model is significantly better. The statistical test is a verification of the results shown in Table~\ref{tab:architecture}. In most cases our model performs significantly better than many of the baselines, and for the EMG, ECG, PPG, and FD-B datasets the performance of the CiTrus (fp) is statistically better than almost all models. Then, in Table~\ref{tab:pretrain-significance},~\ref{tab:multimodal-significance}, and~\ref{tab:sliding-window-significance} we verify the significance of the other three experiments in the main text, and find that in most cases where we see improvements, the new method is also significantly better.
\FloatBarrier
\begin{table*}[ht]
    \begin{tabular}{l|ccc|ccc|ccc|ccc|}
    \hline 
 Dataset & \multicolumn{3}{c}{EMG@20\%} & \multicolumn{3}{c}{ECG@0.5\%} & \multicolumn{3}{c}{PPG@10\%} & \multicolumn{3}{c}{HAR@10\%} \\ 
 \hline 
 Metric & ACC & ROC & PRC & ACC & ROC & PRC & ACC & ROC & PRC & ACC & ROC & PRC \\ 
 \hline 
PatchTST (s) & \textbf{9e-13} & \textbf{9e-13} & \textbf{9e-13} & \textbf{9e-13} & \textbf{9e-13} & \textbf{8e-11} & \textbf{2e-04} & \textbf{4e-04} & \textbf{1e-05} & 1.0 & 1.0 & 0.93 \\ 
PatchTST (p) & \textbf{9e-13} & \textbf{2e-12} & \textbf{9e-13} & \textbf{3e-10} & \textbf{2e-12} & \textbf{2e-10} & \textbf{5e-04} & \textbf{4e-04} & \textbf{4e-05} & \textbf{5e-06} & \textbf{1e-07} & \textbf{2e-09} \\ 
bioFAME (s) & \textbf{9e-13} & \textbf{2e-12} & \textbf{9e-13} & \textbf{2e-12} & \textbf{9e-13} & \textbf{9e-13} & \textbf{7e-05} & \textbf{8e-04} & \textbf{7e-05} & \textbf{2e-12} & \textbf{9e-13} & \textbf{9e-13} \\ 
bioFAME (mp) & \textbf{9e-13} & \textbf{2e-09} & \textbf{3e-10} & \textbf{5e-09} & \textbf{3e-12} & \textbf{9e-12} & \textbf{4e-05} & \textbf{4e-02} & \textbf{7e-03} & \textbf{3e-05} & \textbf{5e-02} & \textbf{2e-03} \\ 
NLPatchTST (s) & \textbf{9e-13} & \textbf{5e-11} & \textbf{9e-13} & \textbf{9e-13} & \textbf{9e-13} & \textbf{4e-11} & \textbf{1e-05} & \textbf{2e-06} & \textbf{4e-08} & 1.0 & 1.0 & 0.98 \\ 
NLPatchTST (p) & \textbf{9e-13} & \textbf{1e-07} & \textbf{4e-08} & \textbf{1e-10} & \textbf{2e-11} & \textbf{2e-10} & \textbf{3e-04} & \textbf{4e-03} & \textbf{1e-04} & 0.64 & 0.79 & 0.14 \\ 
NLPatchTST (mp) & \textbf{3e-08} & \textbf{9e-12} & \textbf{9e-12} & \textbf{3e-10} & \textbf{4e-11} & \textbf{5e-10} & \textbf{7e-05} & \textbf{1e-03} & \textbf{3e-05} & \textbf{2e-06} & \textbf{8e-06} & \textbf{9e-08} \\ 
SimMTM (s) & \textbf{9e-13} & \textbf{9e-13} & \textbf{9e-13} & \textbf{8e-11} & \textbf{7e-08} & \textbf{2e-08} & \textbf{2e-05} & 0.09 & 0.07 & 1.0 & 1.0 & 1.0 \\ 
SimMTM (p) & \textbf{9e-13} & \textbf{9e-13} & \textbf{9e-13} & \textbf{9e-12} & \textbf{2e-09} & \textbf{4e-08} & \textbf{1e-07} & \textbf{2e-03} & \textbf{1e-03} & 1.0 & 1.0 & 1.0 \\ 
Ci (s) & 0.06 & 0.06 & 0.09 & \textbf{2e-10} & \textbf{2e-04} & \textbf{5e-06} & \textbf{6e-04} & \textbf{2e-06} & \textbf{2e-06} & 1.0 & 1.0 & 1.0 \\ 
Ci (p) & \textbf{2e-04} & \textbf{3e-03} & \textbf{2e-03} & \textbf{4e-10} & \textbf{3e-06} & \textbf{2e-05} & \textbf{2e-05} & \textbf{1e-04} & \textbf{3e-06} & 1.0 & 1.0 & 1.0 \\ 
CiTrus (s) & \textbf{2e-02} & 0.38 & 0.51 & \textbf{6e-09} & \textbf{4e-02} & \textbf{4e-03} & \textbf{6e-04} & \textbf{2e-04} & \textbf{3e-03} & 1.0 & 1.0 & 1.0 \\ 
CiTrus (p) & \textbf{4e-06} & \textbf{8e-04} & \textbf{3e-05} & \textbf{3e-09} & \textbf{5e-08} & \textbf{3e-07} & 0.08 & \textbf{1e-02} & \textbf{4e-02} & 1.0 & 1.0 & 1.0 \\ 
CiTrus (mp) & \textbf{5e-06} & \textbf{2e-02} & \textbf{3e-04} & \textbf{1e-07} & \textbf{5e-04} & \textbf{2e-02} & 0.13 & \textbf{2e-02} & \textbf{5e-03} & 1.0 & 1.0 & 1.0 \\ 
\hline 
 Dataset & \multicolumn{3}{c}{FDB@0.5\%} & \multicolumn{3}{c}{Epilepsy@1\%} & \multicolumn{3}{c}{Gesture@10\%} & \multicolumn{3}{c}{SleepInference@0.5\%} \\ 
 \hline 
 Metric & ACC & ROC & PRC & ACC & ROC & PRC & ACC & ROC & PRC & ACC & ROC & PRC \\ 
 \hline 
PatchTST (s) & \textbf{9e-13} & \textbf{9e-13} & \textbf{9e-13} & \textbf{1e-02} & \textbf{2e-05} & \textbf{2e-04} & \textbf{5e-05} & 0.09 & \textbf{1e-04} & 1.0 & 1.0 & 1.0 \\ 
PatchTST (p) & \textbf{3e-12} & \textbf{9e-13} & \textbf{9e-13} & 0.08 & \textbf{7e-06} & \textbf{1e-05} & 0.17 & 0.97 & 0.7 & 1.0 & 1.0 & 1.0 \\ 
bioFAME (s) & \textbf{9e-13} & \textbf{9e-13} & \textbf{9e-13} & \textbf{7e-03} & 0.93 & 1.0 & \textbf{9e-13} & \textbf{4e-10} & \textbf{3e-12} & 1.0 & 1.0 & 1.0 \\ 
bioFAME (mp) & \textbf{3e-12} & \textbf{9e-13} & \textbf{9e-13} & \textbf{6e-03} & 1.0 & 1.0 & \textbf{2e-09} & \textbf{5e-03} & \textbf{3e-08} & 1.0 & 1.0 & 1.0 \\ 
NLPatchTST (s) & \textbf{9e-13} & \textbf{9e-13} & \textbf{9e-13} & \textbf{9e-13} & \textbf{9e-13} & \textbf{9e-13} & \textbf{3e-05} & 0.47 & \textbf{2e-04} & 1.0 & 1.0 & 1.0 \\ 
NLPatchTST (p) & \textbf{6e-12} & \textbf{2e-12} & \textbf{9e-13} & \textbf{7e-04} & \textbf{2e-08} & \textbf{2e-08} & \textbf{5e-02} & 0.98 & 0.7 & 1.0 & 1.0 & 1.0 \\ 
NLPatchTST (mp) & \textbf{2e-11} & \textbf{2e-12} & \textbf{9e-13} & \textbf{7e-03} & \textbf{5e-07} & \textbf{7e-06} & 0.64 & 0.99 & 0.91 & 1.0 & 1.0 & 1.0 \\ 
SimMTM (s) & \textbf{9e-13} & \textbf{2e-12} & \textbf{2e-12} & 0.83 & \textbf{9e-13} & \textbf{9e-13} & 0.83 & 1.0 & 1.0 &  &  &  \\ 
SimMTM (p) & \textbf{2e-08} & \textbf{2e-09} & \textbf{3e-11} & 0.12 & \textbf{9e-13} & \textbf{9e-13} & 1.0 & 1.0 & 1.0 &  &  &  \\ 
Ci (s) & \textbf{5e-06} & \textbf{8e-04} & \textbf{1e-03} & 0.86 & 1.0 & 1.0 & \textbf{4e-08} & \textbf{2e-06} & \textbf{3e-08} & 1.0 & 1.0 & 1.0 \\ 
Ci (p) & \textbf{7e-08} & \textbf{5e-08} & \textbf{2e-09} & 1.0 & 1.0 & 1.0 & \textbf{3e-08} & 0.07 & \textbf{2e-07} & 1.0 & 1.0 & 1.0 \\ 
CiTrus (s) & \textbf{1e-07} & \textbf{3e-07} & \textbf{5e-09} & 0.09 & 1.0 & 1.0 & \textbf{1e-05} & 0.24 & \textbf{2e-03} & 1.0 & 1.0 & 1.0 \\ 
CiTrus (p) & \textbf{5e-07} & \textbf{3e-10} & \textbf{5e-12} & 0.98 & 0.87 & 0.94 & \textbf{2e-06} & \textbf{4e-02} & \textbf{1e-03} & 1.0 & 1.0 & 1.0 \\ 
CiTrus (mp) & \textbf{5e-07} & \textbf{3e-09} & \textbf{1e-10} & 1.0 & 0.35 & 0.16 & \textbf{2e-02} & 0.93 & 0.6 & 1.0 & 1.0 & 1.0 \\ 

    \hline
    \end{tabular}
    \caption{The p-value for the one-sided Wilcoxon signed-rank test across seeds and folds. Significant values are made bold, and indicate that improvements in Table~\ref{tab:architecture} are significant. The comparisons are all done compared to our CiTrus (fp) model, with the one-sided assumption that CiTrus (fp)'s performance is higher.}
\label{tab:architecture-significance}
    \end{table*}
\FloatBarrier

\clearpage
\FloatBarrier
\begin{table*}[ht]
\centering
\begin{tabular}{l|ccc|ccc|ccc|ccc|}
\hline 
 Dataset & \multicolumn{3}{c}{EMG} & \multicolumn{3}{c}{ECG} & \multicolumn{3}{c}{PPG} & \multicolumn{3}{c}{HAR} \\ 
 \hline 
 Data & 20\% & 50\% & 80\% & 0.5\% & 1\% & 2\% & 10\% & 20\% & 50\% & 10\% & 20\% & 50\% \\ 
 \hline 
PatchTST & 0.67 & 0.66 & 0.67 & 0.44 & \textbf{2e-02} & 0.18 & 0.8 & \textbf{4e-02} & 0.98 & \textbf{2e-08} & \textbf{2e-10} & \textbf{2e-04} \\ 
bioFAME & 0.52 & 0.24 & \textbf{8e-04} & \textbf{3e-04} & \textbf{2e-05} & \textbf{3e-08} & 0.92 & 0.17 & 0.3 & \textbf{1e-08} & \textbf{5e-07} & \textbf{9e-12} \\ 
NLPatchTST (p) & 0.17 & \textbf{7e-05} & \textbf{3e-06} & 0.43 & 0.28 & 0.44 & \textbf{2e-03} & 0.45 & 0.78 & \textbf{4e-03} & \textbf{2e-05} & \textbf{3e-02} \\ 
NLPatchTST (mp) & \textbf{3e-04} & \textbf{2e-07} & \textbf{3e-07} & 0.06 & 0.68 & 0.73 & \textbf{3e-03} & 0.84 & 0.19 & \textbf{2e-07} & \textbf{5e-09} & \textbf{5e-05} \\ 
Ci & \textbf{2e-02} & \textbf{3e-06} & 0.35 & 0.36 & 0.08 & 0.64 & 0.85 & 0.3 & 0.07 & \textbf{2e-12} & \textbf{3e-08} & 0.16 \\ 
CiTrus (p) & 0.07 & \textbf{1e-05} & \textbf{8e-06} & 0.05 & 0.56 & 0.45 & \textbf{2e-02} & 0.66 & 0.39 & 0.08 & \textbf{7e-05} & 0.28 \\ 
CiTrus (fp) & \textbf{7e-04} & \textbf{2e-03} & \textbf{4e-02} & \textbf{1e-08} & \textbf{3e-07} & \textbf{5e-08} & \textbf{1e-03} & \textbf{2e-03} & \textbf{8e-03} & \textbf{3e-08} & \textbf{9e-04} & 0.73 \\ 
CiTrus (mp) & 0.77 & \textbf{2e-02} & \textbf{3e-02} & 0.62 & 0.99 & \textbf{1e-02} & \textbf{9e-03} & 0.54 & 0.44 & 0.43 & \textbf{6e-05} & \textbf{5e-03} \\ 
\hline 
 Dataset & \multicolumn{3}{c}{FDB} & \multicolumn{3}{c}{Epilepsy} & \multicolumn{3}{c}{Gesture} & \multicolumn{3}{c}{SleepInference} \\ 
 \hline 
 Data & 0.5\% & 1\% & 2\% & 1\% & 5\% & 10\% & 10\% & 20\% & 50\% & 0.5\% & 1\% & 2\% \\ 
 \hline 
PatchTST & 0.42 & 0.11 & 0.94 & \textbf{4e-02} & 0.75 & \textbf{6e-03} & \textbf{7e-05} & \textbf{2e-05} & \textbf{4e-03} & \textbf{2e-12} & 0.12 & 0.12 \\ 
bioFAME & 0.08 & \textbf{3e-02} & \textbf{3e-05} & 0.43 & 0.77 & \textbf{2e-02} & \textbf{2e-04} & \textbf{2e-04} & \textbf{1e-04} & 0.47 & 0.88 & 0.62 \\ 
NLPatchTST (p) & \textbf{2e-04} & \textbf{2e-05} & \textbf{2e-03} & \textbf{3e-11} & \textbf{5e-11} & \textbf{3e-11} & \textbf{4e-03} & \textbf{3e-03} & \textbf{1e-02} & \textbf{2e-12} & 0.12 & 0.12 \\ 
NLPatchTST (mp) & 0.56 & \textbf{4e-05} & 0.54 & \textbf{1e-10} & \textbf{4e-12} & \textbf{1e-09} & \textbf{3e-06} & \textbf{1e-05} & \textbf{6e-04} & \textbf{2e-12} & 0.12 & 0.12 \\ 
Ci & 0.09 & \textbf{4e-05} & \textbf{1e-04} & \textbf{7e-03} & \textbf{7e-06} & \textbf{1e-06} & \textbf{2e-03} & \textbf{5e-06} & 0.87 & \textbf{4e-04} & 0.88 & 0.12 \\ 
CiTrus (p) & 0.15 & \textbf{4e-02} & \textbf{6e-03} & \textbf{2e-02} & 0.68 & 0.13 & 0.37 & \textbf{1e-04} & 0.7 & \textbf{2e-12} & 0.12 & 0.88 \\ 
CiTrus (fp) & 0.17 & \textbf{2e-03} & \textbf{1e-05} & 0.18 & \textbf{8e-03} & 0.3 & \textbf{2e-05} & 0.25 & \textbf{9e-03} & \textbf{2e-12} & 0.12 & 0.12 \\ 
CiTrus (mp) & 0.52 & 0.42 & 0.44 & \textbf{3e-05} & 0.12 & 0.14 & \textbf{2e-04} & \textbf{2e-02} & 0.95 & 0.18 & 0.12 & 0.62 \\ 

\hline
\end{tabular}
\caption{The p-value for the one-sided Wilcoxon signed-rank test for accuracy values across seeds and folds. Significant values are made bold, and indicate that improvements in Table~\ref{tab:pretraining} are significant.}
\label{tab:pretrain-significance}
\end{table*}
\FloatBarrier

\FloatBarrier
\begin{table*}[ht]
\centering
\begin{tabular}{l|ccc|ccc|ccc|ccc|}
\hline 
 Dataset & \multicolumn{3}{c}{EMG} & \multicolumn{3}{c}{ECG} & \multicolumn{3}{c}{PPG} & \multicolumn{3}{c}{HAR} \\ 
 \hline 
 Data & 20\% & 50\% & 80\% & 0.5\% & 1\% & 2\% & 10\% & 20\% & 50\% & 10\% & 20\% & 50\% \\ 
 \hline 
NLPatchTST & \textbf{2e-02} & 0.12 & 0.21 & \textbf{3e-02} & 0.17 & 0.21 & 0.55 & 0.42 & 0.1 & \textbf{6e-06} & \textbf{3e-04} & \textbf{8e-04} \\ 
CiTrus & 0.06 & \textbf{3e-03} & \textbf{3e-03} & 0.23 & 0.64 & \textbf{4e-02} & 0.96 & 0.51 & 0.72 & 0.13 & 0.26 & \textbf{4e-02} \\ 
\hline 
 Dataset & \multicolumn{3}{c}{FDB} & \multicolumn{3}{c}{Epilepsy} & \multicolumn{3}{c}{Gesture} & \multicolumn{3}{c}{SleepInference} \\ 
 \hline 
 Data & 0.5\% & 1\% & 2\% & 1\% & 5\% & 10\% & 10\% & 20\% & 50\% & 0.5\% & 1\% & 2\% \\ 
 \hline 
NLPatchTST & \textbf{3e-03} & 0.88 & \textbf{4e-03} & 0.28 & 0.15 & \textbf{5e-03} & \textbf{3e-03} & 0.09 & 0.5 & \textbf{2e-12} & 0.12 & 0.12 \\ 
CiTrus & 0.45 & 0.36 & 0.16 & 0.14 & 0.38 & 0.69 & \textbf{9e-06} & \textbf{9e-07} & 0.81 & \textbf{4e-12} & 0.38 & 0.25 \\ 

\hline
\end{tabular}
\caption{The p-value for the one-sided Wilcoxon signed-rank test for accuracy values across seeds and folds. Significant values are made bold, and indicate that improvements in Table~\ref{tab:multimodal} are significant.}
\label{tab:multimodal-significance}
\end{table*}
\FloatBarrier

\clearpage
\FloatBarrier
\begin{table*}[ht]
\centering
\begin{tabular}{l|ccc|ccc|ccc|ccc|}
\hline 
 Dataset & \multicolumn{3}{c}{ECG} & \multicolumn{3}{c}{PPG} & \multicolumn{3}{c}{HAR} & \multicolumn{3}{c}{Gesture} \\ 
 \hline 
 Data & 0.5\% & 1\% & 2\% & 10\% & 20\% & 50\% & 10\% & 20\% & 50\% & 10\% & 20\% & 50\% \\ 
 \hline 
PatchTST (s) & 0.53 & 0.31 & 0.12 & \textbf{2e-12} & \textbf{2e-10} & \textbf{2e-12} & \textbf{2e-12} & \textbf{2e-12} & \textbf{2e-12} & \textbf{5e-04} & \textbf{5e-07} & \textbf{5e-05} \\ 
PatchTST (p) & 0.07 & \textbf{1e-03} & 0.07 & \textbf{2e-12} & \textbf{2e-10} & \textbf{2e-12} & \textbf{3e-11} & \textbf{5e-08} & \textbf{2e-12} & 0.06 & \textbf{3e-03} & \textbf{2e-02} \\ 
bioFAME (s) & \textbf{2e-02} & \textbf{3e-02} & \textbf{6e-03} & \textbf{2e-12} & \textbf{5e-12} & \textbf{2e-12} & \textbf{2e-12} & \textbf{5e-12} & \textbf{2e-12} & 0.06 & \textbf{5e-05} & \textbf{3e-04} \\ 
bioFAME (mp) & \textbf{3e-08} & \textbf{6e-11} & \textbf{2e-11} & \textbf{2e-12} & \textbf{1e-07} & \textbf{2e-12} & \textbf{9e-12} & \textbf{2e-12} & \textbf{2e-12} & 0.37 & 0.17 & 0.69 \\ 
SimMTM (s) & 0.2 & \textbf{9e-09} & \textbf{1e-03} & \textbf{1e-06} & \textbf{1e-06} & \textbf{2e-07} & 0.83 & 0.23 & 0.69 & \textbf{1e-04} & 0.97 & 0.09 \\ 
SimMTM (p) & 0.52 & 0.11 & 0.5 & \textbf{3e-03} & \textbf{1e-05} & \textbf{3e-07} & \textbf{2e-02} & 0.12 & 0.4 & 0.29 & 0.37 & 0.35 \\ 
NLPatchTST (s) & 0.93 & 0.3 & 0.15 & \textbf{8e-08} & \textbf{3e-07} & \textbf{2e-12} & \textbf{4e-12} & \textbf{2e-12} & \textbf{2e-12} & \textbf{5e-02} & \textbf{3e-06} & \textbf{3e-02} \\ 
NLPatchTST (p) & 0.99 & 0.29 & 0.13 & \textbf{2e-12} & \textbf{1e-11} & \textbf{2e-12} & \textbf{2e-12} & \textbf{2e-12} & \textbf{2e-12} & 0.68 & \textbf{1e-03} & \textbf{5e-03} \\ 
NLPatchTST (mp) & 0.06 & 0.11 & 0.52 & \textbf{2e-12} & \textbf{1e-10} & \textbf{2e-12} & \textbf{1e-10} & \textbf{9e-12} & \textbf{2e-12} & \textbf{6e-03} & \textbf{3e-03} & \textbf{8e-04} \\ 
Ci (s) & 0.14 & 0.24 & 0.37 & \textbf{5e-08} & \textbf{2e-07} & \textbf{4e-06} & \textbf{2e-12} & \textbf{2e-12} & \textbf{2e-12} & 0.72 & \textbf{3e-03} & 0.16 \\ 
Ci (p) & 0.48 & 0.39 & 0.48 & \textbf{2e-12} & \textbf{4e-10} & \textbf{2e-12} & \textbf{4e-12} & \textbf{2e-12} & \textbf{2e-12} & 0.29 & 0.33 & \textbf{2e-03} \\ 
CiTrus (s) & 0.95 & 0.11 & 0.6 & \textbf{3e-11} & \textbf{2e-07} & \textbf{4e-08} & \textbf{2e-12} & \textbf{2e-12} & \textbf{2e-12} & \textbf{1e-03} & \textbf{2e-04} & 0.08 \\ 
CiTrus (p) & 0.4 & 0.2 & 0.13 & \textbf{2e-12} & \textbf{9e-12} & \textbf{4e-12} & \textbf{2e-12} & \textbf{2e-12} & \textbf{2e-12} & 0.56 & 0.98 & 0.21 \\ 
CiTrus (fp) & 0.57 & 0.69 & \textbf{2e-02} & \textbf{2e-12} & \textbf{2e-12} & \textbf{7e-08} & \textbf{2e-12} & \textbf{4e-12} & \textbf{2e-12} & \textbf{7e-04} & \textbf{3e-02} & \textbf{1e-02} \\ 
CiTrus (mp) & 0.07 & 0.32 & \textbf{2e-02} & \textbf{2e-12} & \textbf{7e-10} & \textbf{2e-12} & \textbf{2e-12} & \textbf{2e-12} & \textbf{2e-12} & 0.23 & \textbf{2e-05} & 0.8 \\ 

\hline
\end{tabular}
\caption{The p-value for the double-sided Wilcoxon signed-rank test for accuracy values across seeds and folds. Significant values are made bold, and indicate that improvements in Table~\ref{tab:sliding-window} are significant.}
\label{tab:sliding-window-significance}
\end{table*}
\FloatBarrier

\clearpage
\section{2 second or 30 second pre-training}
\label{app:time-length}
The following table shows the percentage improvement of using $30$ second pre-training windows for the SleepEDF dataset, and interpolating the size of the dataset to $3000$ timesteps. Using $30$s windows for pre-training is described in the bioFAME paper~\cite{liu2023frequency}, whereas using $2$s windows and resampling the fine-tuning datasets to $200$ timesteps is described in the TFC paper~\cite{zhang2022self}. Clearly, for the FD-B and EMG datasets, using longer pre-training data and resampling to those longer pre-training inputs improves performance. The same is true for PPG, although the improvements are less than using our sliding-window approach, as shown in Table~\ref{tab:sliding-window}. Although the long pre-training and resampling is sometimes better for the ECG datasets, this is only the case for model architectures that are outperformed by the other model architectures, as shown in Table~\ref{tab:architecture}, for example. For the Gesture and HAR datasets the long pre-training and resampling does not seem to improve performance, and in many cases even slightly degrades performance.
\FloatBarrier
\begin{table*}[ht]
\centering
\begin{tabular}{l|ccc|ccc|ccc|ccc|}
\hline 
 Dataset & \multicolumn{3}{c}{ECG} & \multicolumn{3}{c}{PPG} & \multicolumn{3}{c}{HAR} \\ 
 \hline 
 Data & 0.5\% & 1\% & 2\% & 10\% & 20\% & 50\% & 10\% & 20\% & 50\% \\ 
 \hline 
PatchTST (s) & \textbf{+15.1} & \textbf{+3.2} & \textbf{+1.7} & \textbf{+42.8} & \textbf{+51.5} & \textbf{+50.6} & \textbf{+2.1} & \textbf{+2.1} & \textbf{+5.3} \\ 
PatchTST (p) & \textbf{+18.1} & \textbf{+19.4} & \textbf{+10.7} & \textbf{+22.5} & \textbf{+42.7} & \textbf{+46.2} & \textbf{+10.2} & \textbf{+10.2} & \textbf{+10.4} \\ 
bioFAME (s) & \textbf{+12.2} & \textbf{+7.2} & \textbf{+8.1} & \textbf{+59.3} & \textbf{+56.7} & \textbf{+47.0} & \textbf{+2.4} & -6.4 & -6.0 \\ 
bioFAME (mp) & \textbf{+32.6} & \textbf{+21.2} & \textbf{+16.3} & \textbf{+51.1} & \textbf{+52.4} & \textbf{+60.0} & \textbf{+8.9} & -3.6 & \textbf{+12.5} \\ 
NLPatchTST (s) & \textbf{+2.0} & -6.1 & -3.2 & \textbf{+33.1} & \textbf{+36.2} & \textbf{+50.4} & -0.5 & \textbf{+2.2} & \textbf{+4.7} \\ 
NLPatchTST (p) & \textbf{+14.1} & \textbf{+15.9} & \textbf{+12.4} & \textbf{+35.0} & \textbf{+42.1} & \textbf{+60.7} & \textbf{+3.8} & \textbf{+6.5} & \textbf{+1.9} \\ 
NLPatchTST (mp) & \textbf{+22.1} & \textbf{+24.7} & \textbf{+14.1} & \textbf{+40.5} & \textbf{+47.6} & \textbf{+60.4} & \textbf{+4.0} & \textbf{+6.9} & \textbf{+5.6} \\ 
Ci (s) & -3.3 & -2.6 & -3.0 & \textbf{+20.3} & \textbf{+31.0} & \textbf{+17.4} & -7.7 & -3.9 & -2.1 \\ 
Ci (p) & -1.5 & -1.9 & -2.9 & \textbf{+38.2} & \textbf{+40.1} & \textbf{+32.1} & -6.9 & -6.4 & -8.8 \\ 
CiTrus (s) & -6.2 & -2.3 & -6.1 & \textbf{+19.0} & \textbf{+21.9} & \textbf{+17.6} & -5.6 & \textbf{+1.9} & \textbf{+4.0} \\ 
CiTrus (p) & \textbf{+5.5} & \textbf{+6.6} & \textbf{+5.6} & \textbf{+36.5} & \textbf{+44.8} & \textbf{+45.8} & -4.8 & -6.2 & -4.7 \\ 
CiTrus (fp) & -15.6 & -12.5 & -14.1 & \textbf{+6.6} & \textbf{+6.6} & \textbf{+2.7} & -21.0 & -24.2 & -23.5 \\ 
CiTrus (mp) & \textbf{+14.9} & \textbf{+8.0} & \textbf{+6.6} & \textbf{+39.5} & \textbf{+40.6} & \textbf{+45.8} & -5.6 & -5.8 & -2.8 \\ 
\hline 
 Dataset & \multicolumn{3}{c}{Gesture} & \multicolumn{3}{c}{FDB} & \multicolumn{3}{c}{EMG} \\ 
 \hline 
 Data & 10\% & 20\% & 50\% & 0.5\% & 1\% & 2\% & 20\% & 50\% & 80\% \\ 
 \hline 
PatchTST (s) & -18.4 & -17.9 & -9.6 & \textbf{+9.9} & \textbf{+20.8} & \textbf{+15.5} & \textbf{+85.7} & \textbf{+22.0} & \textbf{+6.0} \\ 
PatchTST (p) & -7.9 & -9.0 & -6.1 & \textbf{+11.1} & \textbf{+17.7} & \textbf{+16.5} & \textbf{+60.6} & \textbf{+43.7} & \textbf{+20.4} \\ 
bioFAME (s) & \textbf{+33.5} & \textbf{+22.1} & \textbf{+18.5} & \textbf{+24.0} & \textbf{+27.2} & \textbf{+23.0} & \textbf{+70.9} & \textbf{+34.4} & \textbf{+14.7} \\ 
bioFAME (mp) & \textbf{+0.9} & \textbf{+4.6} & \textbf{+6.7} & \textbf{+21.7} & \textbf{+22.0} & \textbf{+14.9} & \textbf{+27.1} & \textbf{+15.3} & \textbf{+10.8} \\ 
NLPatchTST (s) & -21.7 & -20.9 & -10.3 & \textbf{+13.4} & \textbf{+19.3} & \textbf{+13.2} & \textbf{+41.3} & \textbf{+10.1} & \textbf{+3.6} \\ 
NLPatchTST (p) & -9.1 & -8.7 & -3.7 & \textbf{+21.7} & \textbf{+24.7} & \textbf{+23.3} & \textbf{+28.4} & \textbf{+8.5} & \textbf{+3.6} \\ 
NLPatchTST (mp) & -10.3 & -5.7 & -4.0 & \textbf{+14.0} & \textbf{+16.8} & \textbf{+15.3} & \textbf{+22.7} & \textbf{+10.1} & \textbf{+5.4} \\ 
Ci (s) & -21.7 & -35.3 & -10.1 & \textbf{+7.6} & \textbf{+4.2} & \textbf{+4.2} & \textbf{+1.7} & \textbf{+1.1} & \textbf{+0.4} \\ 
Ci (p) & -29.2 & -28.6 & -9.2 & \textbf{+9.0} & \textbf{+6.3} & \textbf{+3.5} & \textbf{+2.9} & \textbf{+1.4} & \textbf{+0.6} \\ 
CiTrus (s) & -18.7 & -23.6 & -17.8 & \textbf{+12.1} & \textbf{+6.9} & \textbf{+5.4} & \textbf{+0.3} & \textbf{+0.2} & -0.1 \\ 
CiTrus (p) & \textbf{+0.7} & -1.7 & -5.6 & \textbf{+1.0} & \textbf{+4.7} & \textbf{+3.6} & \textbf{+5.5} & \textbf{+0.3} & \textbf{+0.0} \\ 
CiTrus (fp) & -35.0 & -38.1 & -29.2 & -5.1 & -3.2 & -0.8 & -7.2 & -9.8 & -6.3 \\ 
CiTrus (mp) & -4.1 & -1.3 & -2.0 & \textbf{+6.9} & \textbf{+7.4} & \textbf{+5.6} & \textbf{+2.9} & \textbf{+2.3} & \textbf{+0.6} \\ 

\hline
\end{tabular}
\caption{A comparison between using $2$s windows from the SleepEDF pre-training dataset and resampling the fine-tuning dataset to $200$ timesteps as described in~\cite{zhang2022self}, and using $30$s windows from the SleepEDF pre-training dataset, as described in~\cite{liu2023frequency}, and resampling the fine-tuning dataset to $3000$ timesteps. Each value indicates the average performance improvement (across all three metrics) in percentages. Values larger than $0$ are made bold.}
\label{tab:long-pretraining}
\end{table*}
\FloatBarrier

\clearpage
\section{Exact data parameters}
\label{app:data}
The Epilepsy, Gesture, EMG, and FD-B datasets are from the TFC paper~\cite{zhang2022self}. \begin{links}\link{The link to their code:}{https://github.com/mims-harvard/TFC-pretraining/tree/main}\end{links}.

To download and pre-process the ECG, HAR, and PPG datasets, we based our code on~\cite{xu2023retrieval}.
\begin{links}\link{The link to their code:}{https://github.com/maxxu05/rebar}\end{links}.

The script to download and pre-process the specific subset of the SleepEDF data is based on~\cite{eldele2023self}. \begin{links}\link{The link to their code:}{https://gist.github.com/emadeldeen24/a22691e36759934e53984289a94cb09b}\end{links}.

\paragraph{SleepEDF}
The SleepEDF dataset~\cite{eldele2023self} contains 197 whole-night sleep recordings. It records two EEG channels (Fpz-Cz and Pz-Oz), an EOG channel, a respiratory (oro-nasal), an EMG channel located below the skin, and rectal temperature during each night of sleep. Similar to~\cite{eldele2023self,liu2023frequency}, we use the cassette study, which is a subset of the dataset, that focuses on healthy caucasian subjects. Each $30$ seconds in the final, preprocessed dataset, is associated with one of five sleeping stages: Rapid Eye Movement (REM), Non-rapid eye movement (N1, N2, N3), and Wake (W).

\paragraph{Epilepsy}
The Epilepsy dataset~\cite{andrzejak2001indications} contains single-channel recordings from $500$ subjects. Each time window that is labeled corresponds to $1$ second, sampled at $178$Hz, and is thus $178$ timesteps long. In total, there are $11,500$ samples with two classes: whether a subject is having a seizure in that $1$-second time window or not.

\paragraph{Gesture}
The gesture dataset~\cite{liu2009uwave} contains 3-dimensional accelerometer data that exemplify $8$ different hand gestures. Each time window corresponds to a specific gesture, and is $2.06$seconds long. Although the sampling frequency is not clear~\cite{zhang2022self}, it is assumed to be $100$Hz, which is a common sampling frequency for accelerometer datasets like this. This means each sample is $206$ timesteps long. The eight gestures in the datasets are: swiping left, swiping right, swiping up, swiping down, waving counterclockwise, waving clockwise, waving in a square, and waving in a right arrow.

\paragraph{EMG}
The EMG dataset~\cite{goldberger2000physiobank} contains data from three different patients, a patient with neuropathy, myopathy, and a healthy volunteer. The timeseries from each subject is split into $1500$ timestep windows, sampled at $4$kHz the time windows correspond to $375$ms of data. The label for each time window is the subject it came from. The goal is thus to classify whether the observed EMG in a specific time window is from someone suffering from neuropathy, myopathy, or a healthy volunteer.

\paragraph{FD-B}
The FD-B dataset~\cite{lessmeier2016condition} is a fault detection dataset generated by an electromechanical drive system that monitors the rolling bearings. Each time window corresponds to a different fault in the electromotor. Specifically, the three classes are that a rolling is undamaged, inner damaged, and outer damaged. The data has a single channel, and is separated into time windows using a $5120$ timestep sliding window, with a sampling frequency of $64$kHz, each time window corresponds to $80$ms.

\paragraph{ECG}
The ECG dataset~\cite{moody1983new} contains dual channel electrocardiogram recordings, where each $10$-second time-window, with a $250$Hz sampling frequency, corresponds to $2500$ time points. Each time window is either labeled as atrial fibrillation or a normal heart rhythm.

\paragraph{PPG}
The PPG dataset~\cite{schmidt2018introducing} contains single-channel photoplethysmogram recordings to perform affect detection. Each time window is one minute long, with a $64$Hz sampling frequency, this corresponds to $3840$ time steps. The possible labels for each time window are baseline, stress, amusement, and meditation.

\paragraph{HAR}
The HAR dataset~\cite{reyes2015smartphone} contains 3-dimensional accelerometer and gyroscopic sensor data to classify daily activities. The total number of channels in the data is $6$, and each labeled time window is $2.56$s long, with a $50$Hz sampling frequency, this corresponds to $128$ timesteps. There are $6$ labels; walking, walking upstairs, walking downstairs, sitting, standing, and laying.

\clearpage
\section{Exact model parameters}
\label{app:model}
\paragraph{Residual blocks in the convolutional encoder}
Each layer in the convolutional encoder is a residual block.
One path through the residual block is a single convolution layer, with a kernel size of $3$, a stride of $2$, and a padding of $1$, that increases the number of input channels C to $2 *$C.
The other path are two sequential convolutional layers, both with a kernel size of $3$.
The first convolutional layer has a stride of $1$, and a padding of $1$, without the bias parameters, and increases the number of input channels C to $2 * $C.
The second convolutional layer has a stride of $2$, and a padding of $1$, also without the bias parameters, and does not increase the number of channels further.
The first convolutional layer is followed by 1D BatchNorm, a GELU activation, Dropout, and then the second convolutional layer.
The second convolutional layer is followed by 1D BatchNorm, a GELU activation, and its output is added to the other path.
The sum of the two paths then goes through a GELU activation and Dropout.

\paragraph{Additional parameters}
Throughout each of the models we use a $0.1$ dropout for both pre-training, and fine-tuning. Moreover, during the fine-tuning stage, when the pre-train head is replaced with the linear classification layer, we add a $0.5$ Dropout layer before the linear classification layer.
This is to reduce overfitting on the (often small) fine-tuning datasets.

\clearpage
\section{Middle and high data regime results}
\label{app:full-results}
Given the large size of the tables that display the result, we decided to only show the lowest data regime (the smallest percentage of training + validation available) in the main paper, see Table~\ref{tab:architecture}. In the following two subsections, we show the middle and high regime results for each of the datasets and each of the models. The models are trained the same way as outlined in the main text.
\subsection{Middle-data regime}
\FloatBarrier
\begin{table*}[ht]
    \begin{tabular}{l|ccc|ccc|ccc|ccc|}
    \hline 
 Dataset & \multicolumn{3}{c}{EMG@50\%} & \multicolumn{3}{c}{ECG@1\%} & \multicolumn{3}{c}{PPG@20\%} & \multicolumn{3}{c}{HAR@20\%} \\ 
 \hline 
 Metric & ACC & ROC & PRC & ACC & ROC & PRC & ACC & ROC & PRC & ACC & ROC & PRC \\ 
 \hline 
PatchTST (s) & 78.37 & 89.91 & 84.03 & 56.34 & 52.92 & 61.79 & 57.07 & 73.92 & 53.98 & 75.11 & 91.05 & 76.52 \\ 
PatchTST (p) & 65.32 & 80.0 & 70.9 & 60.04 & 53.23 & 61.86 & 58.73 & 73.8 & 54.56 & 68.27 & 88.89 & 70.99 \\ 
bioFAME (s) & 63.95 & 78.8 & 70.97 & 63.39 & 52.46 & 61.21 & 58.62 & 74.99 & 55.85 & 59.25 & 86.77 & 63.77 \\ 
bioFAME (mp) & 78.67 & 93.8 & 88.33 & 69.78 & 53.35 & 61.85 & 57.62 & 75.01 & 55.48 & 65.81 & 88.73 & 70.19 \\ 
NLPatchTST (s) & 85.77 & 95.29 & 90.9 & 58.5 & 52.76 & 61.35 & 58.21 & 74.08 & 54.43 & 76.27 & 91.49 & 77.89 \\ 
NLPatchTST (p) & 86.89 & 96.04 & 92.45 & 56.99 & 52.89 & 61.47 & 59.22 & 74.67 & 55.54 & 72.13 & 90.23 & 74.12 \\ 
NLPatchTST (mp) & 86.01 & 95.3 & 91.5 & 59.01 & 53.26 & 61.75 & 58.43 & 74.48 & 55.22 & 68.28 & 88.76 & 70.8 \\ 
SimMTM (s) & 90.21 & 97.0 & 93.12 & \doubleunderline{77.76} & 55.12 & 63.48 & 55.26 & \doubleunderline{77.72} & \doubleunderline{58.74} & 78.85 & 92.02 & 81.34 \\ 
SimMTM (p) & 86.05 & 94.01 & 90.59 & 62.82 & 54.01 & 62.32 & 54.76 & \underline{76.98} & 58.31 & 79.24 & 92.28 & 81.99 \\ 
Ci (s) & \textbf{98.16} & 99.1 & 97.77 & 73.28 & 55.74 & 64.24 & \underline{60.08} & 75.61 & 57.11 & \textbf{84.15} & \textbf{93.86} & \textbf{85.6} \\ 
Ci (p) & 96.37 & 99.26 & 98.42 & 70.58 & 55.27 & 63.73 & 58.94 & 75.86 & 57.23 & 78.95 & 92.72 & 82.13 \\ 
CiTrus (s) & \doubleunderline{98.15} & \textbf{99.58} & \textbf{99.72} & \underline{73.36} & \underline{55.95} & \underline{64.4} & 59.67 & 74.15 & 55.26 & 79.14 & 92.19 & 80.87 \\ 
CiTrus (p) & \underline{97.29} & \doubleunderline{99.4} & \underline{98.54} & 71.49 & 55.61 & 64.18 & 58.93 & 75.7 & \underline{58.42} & \underline{82.06} & \underline{92.87} & \underline{82.64} \\ 
CiTrus (fp) & 97.05 & \underline{99.28} & \doubleunderline{98.55} & \textbf{86.76} & \textbf{56.48} & \textbf{64.96} & \textbf{63.9} & \textbf{81.65} & \textbf{65.27} & 75.13 & 90.38 & 76.18 \\ 
CiTrus (mp) & 96.08 & 98.52 & 97.09 & 72.94 & \doubleunderline{55.96} & \doubleunderline{64.66} & \doubleunderline{60.41} & 75.62 & 58.2 & \doubleunderline{82.17} & \doubleunderline{92.96} & \doubleunderline{82.79} \\ 
\hline 
 Dataset & \multicolumn{3}{c}{FDB@1\%} & \multicolumn{3}{c}{Epilepsy@5\%} & \multicolumn{3}{c}{Gesture@20\%} & \multicolumn{3}{c}{SleepEDF@1\%} \\ 
 \hline 
 Metric & ACC & ROC & PRC & ACC & ROC & PRC & ACC & ROC & PRC & ACC & ROC & PRC \\ 
 \hline 
PatchTST (s) & 55.04 & 55.23 & 38.86 & 95.49 & 97.89 & 99.28 & 52.59 & 86.18 & 59.74 & 69.8 & 68.03 & 36.41 \\ 
PatchTST (p) & 62.18 & 56.8 & 40.29 & 95.42 & 97.67 & 99.12 & 58.17 & 88.24 & 65.88 & 71.01 & 68.03 & 36.08 \\ 
bioFAME (s) & 53.18 & 54.4 & 37.97 & 95.02 & 97.93 & 99.38 & 39.51 & 81.63 & 48.31 & 71.08 & 69.29 & 38.74 \\ 
bioFAME (mp) & 58.96 & 58.04 & 41.14 & 95.03 & 97.97 & 99.37 & 45.67 & 83.9 & 54.13 & 71.26 & 69.19 & \underline{39.37} \\ 
NLPatchTST (s) & 63.55 & 58.05 & 41.15 & 85.58 & 81.28 & 90.22 & 54.02 & 86.33 & 59.9 & 68.29 & 67.84 & 36.26 \\ 
NLPatchTST (p) & 66.04 & 58.12 & 41.31 & 94.64 & 96.21 & 98.41 & 57.86 & 88.05 & 66.14 & 72.81 & 68.8 & 38.07 \\ 
NLPatchTST (mp) & 68.04 & 58.96 & 41.88 & 94.97 & 96.83 & 98.72 & \underline{59.38} & \underline{88.28} & \underline{67.35} & 71.16 & 68.01 & 36.5 \\ 
SimMTM (s) & 64.03 & 60.88 & 44.55 & 95.45 & 55.31 & 87.07 & \doubleunderline{64.46} & \doubleunderline{89.55} & \doubleunderline{71.81} &  &  &  \\ 
SimMTM (p) & 80.71 & 66.3 & 50.64 & 94.56 & 55.28 & 87.27 & \textbf{65.71} & \textbf{90.06} & \textbf{72.07} &  &  &  \\ 
Ci (s) & \textbf{84.27} & \textbf{68.14} & \textbf{54.51} & 95.56 & \doubleunderline{98.71} & \doubleunderline{99.58} & 57.01 & 88.07 & 63.54 & \doubleunderline{74.43} & \textbf{70.29} & \textbf{40.24} \\ 
Ci (p) & \doubleunderline{82.03} & \underline{67.55} & \doubleunderline{53.42} & \textbf{96.15} & \textbf{98.83} & \textbf{99.63} & 50.8 & 86.38 & 58.81 & \textbf{75.12} & \doubleunderline{69.87} & \doubleunderline{40.12} \\ 
CiTrus (s) & 80.6 & 67.26 & 52.9 & 95.46 & \underline{98.52} & \underline{99.51} & 56.61 & 88.15 & 63.86 & \underline{74.19} & \underline{69.4} & 38.41 \\ 
CiTrus (p) & 79.13 & 65.33 & 49.96 & 95.58 & 96.99 & 98.61 & 50.18 & 86.51 & 59.77 & 72.42 & 69.03 & 38.37 \\ 
CiTrus (fp) & \underline{82.01} & \doubleunderline{67.61} & \underline{53.21} & \doubleunderline{95.88} & 98.44 & 99.47 & 58.08 & 86.67 & 64.88 & 55.56 & 62.35 & 30.13 \\ 
CiTrus (mp) & 76.37 & 64.97 & 49.52 & \underline{95.76} & 97.61 & 98.94 & 59.2 & 87.82 & 66.14 & 70.46 & 68.07 & 36.46 \\ 
\hline
\end{tabular}
\caption{A comparison of the different model architectures for the middle-data regime. We use the same evaluation method for each model, for EMG and FD-B we interpolate the data for fine-tuning, and for the other datasets we use our sliding window approach. The letters in brackets refer to how the model is trained; (s) is trained from scratch, (p) means it is pre-trained, (mp) means it uses multi-modal pre-training, and (fp) means it uses frequency pre-training. ACC, ROC, and PRC refer to the accuracy, area under the receiver operating characteristic, and the area under the precision recall curve, respectively. The best result for each metric is shown in bold, the second best result is double-underlined, and the third best result is single-underlined.}
\label{tab:middle-data}
\end{table*}
\FloatBarrier

\clearpage
\subsection{High-data regime}
\FloatBarrier
\begin{table*}[ht]
    \begin{tabular}{l|ccc|ccc|ccc|ccc|}
    \hline 
 Dataset & \multicolumn{3}{c}{EMG@80\%} & \multicolumn{3}{c}{ECG@2\%} & \multicolumn{3}{c}{PPG@50\%} & \multicolumn{3}{c}{HAR@50\%} \\ 
 \hline 
 Metric & ACC & ROC & PRC & ACC & ROC & PRC & ACC & ROC & PRC & ACC & ROC & PRC \\ 
 \hline 
PatchTST (s) & 89.4 & 97.59 & 95.18 & 59.57 & 53.44 & 61.97 & 62.34 & 74.82 & 56.08 & 84.3 & 93.52 & 84.22 \\ 
PatchTST (p) & 77.95 & 90.03 & 82.66 & 62.53 & 53.54 & 61.93 & 62.18 & 75.65 & 56.9 & 81.04 & 92.71 & 81.84 \\ 
bioFAME (s) & 76.89 & 90.03 & 81.42 & 67.24 & 52.94 & 61.49 & 60.91 & 76.57 & 55.91 & 66.44 & 88.43 & 69.51 \\ 
bioFAME (mp) & 84.31 & 95.36 & 90.76 & 74.92 & 54.07 & 62.41 & 62.44 & 76.11 & 57.43 & 81.35 & 92.81 & 81.64 \\ 
NLPatchTST (s) & 92.77 & 98.65 & 96.8 & 60.73 & 53.72 & 61.85 & 62.02 & 76.61 & 58.02 & 87.06 & 94.04 & 85.88 \\ 
NLPatchTST (p) & 93.29 & 98.32 & 96.57 & 59.7 & 53.18 & 61.58 & 61.37 & 75.36 & 56.8 & 85.77 & 93.77 & 85.19 \\ 
NLPatchTST (mp) & 91.21 & 97.72 & 95.08 & 61.14 & 53.45 & 61.77 & 62.6 & 75.74 & 57.15 & 84.59 & 93.5 & 84.28 \\ 
SimMTM (s) & 93.05 & 99.04 & 97.75 & 68.11 & 55.02 & 63.49 & 59.59 & \doubleunderline{81.11} & \doubleunderline{64.61} & 82.12 & 92.72 & 83.01 \\ 
SimMTM (p) & 92.45 & 98.73 & 97.49 & 62.29 & 53.87 & 62.24 & 59.26 & 79.11 & \underline{61.72} & 82.01 & 93.0 & 83.81 \\ 
Ci (s) & \textbf{98.88} & \underline{99.52} & 98.98 & 74.44 & \doubleunderline{56.58} & \textbf{65.43} & 62.01 & \underline{79.17} & 61.34 & 87.55 & 94.28 & \underline{86.96} \\ 
Ci (p) & \underline{98.27} & 99.43 & 98.89 & 75.67 & 55.97 & 64.13 & \doubleunderline{63.8} & 78.05 & 60.21 & 86.76 & \doubleunderline{94.34} & 86.89 \\ 
CiTrus (s) & 97.66 & \textbf{99.69} & \textbf{99.76} & \underline{75.77} & 56.18 & 64.72 & \underline{62.9} & 77.67 & 60.21 & \textbf{89.82} & \textbf{94.51} & \textbf{87.71} \\ 
CiTrus (p) & 97.77 & 99.43 & \underline{99.26} & 75.5 & 56.17 & \underline{64.92} & 61.48 & 77.5 & 59.26 & \underline{89.29} & 94.25 & 86.87 \\ 
CiTrus (fp) & \doubleunderline{98.64} & \doubleunderline{99.66} & \doubleunderline{99.69} & \textbf{88.79} & \textbf{56.6} & \doubleunderline{65.15} & \textbf{67.16} & \textbf{83.21} & \textbf{65.68} & \doubleunderline{89.74} & \underline{94.32} & \doubleunderline{87.2} \\ 
CiTrus (mp) & 97.92 & 99.51 & 98.99 & \doubleunderline{79.05} & \underline{56.24} & 64.78 & 61.36 & 76.76 & 57.98 & 88.12 & 94.08 & 86.4 \\ 
\hline 
 Dataset & \multicolumn{3}{c}{FDB@2\%} & \multicolumn{3}{c}{Epilepsy@10\%} & \multicolumn{3}{c}{Gesture@50\%} & \multicolumn{3}{c}{SleepEDF@2\%} \\ 
 \hline 
 Metric & ACC & ROC & PRC & ACC & ROC & PRC & ACC & ROC & PRC & ACC & ROC & PRC \\ 
 \hline 
PatchTST (s) & 67.8 & 60.64 & 43.76 & 95.94 & 98.47 & 99.5 & 60.98 & 89.8 & 68.77 & 70.71 & 68.96 & 38.42 \\ 
PatchTST (p) & 67.58 & 59.24 & 42.26 & 96.25 & 98.32 & 99.36 & 64.29 & 91.45 & 72.91 & 73.02 & 69.12 & 38.56 \\ 
bioFAME (s) & 60.19 & 58.17 & 41.42 & 95.52 & 98.33 & 99.5 & 45.49 & 86.69 & 55.82 & 71.72 & 69.61 & 39.7 \\ 
bioFAME (mp) & 65.26 & 61.77 & 45.46 & 95.77 & 98.43 & 99.51 & 52.1 & 88.16 & 61.26 & 70.5 & 69.02 & 39.33 \\ 
NLPatchTST (s) & 73.81 & 63.47 & 46.63 & 92.01 & 93.02 & 96.46 & 61.56 & 89.31 & 68.34 & 67.69 & 67.45 & 36.42 \\ 
NLPatchTST (p) & 72.29 & 61.52 & 44.37 & 95.91 & 98.21 & 99.39 & 64.46 & 91.23 & 72.09 & \doubleunderline{75.52} & 69.4 & 38.83 \\ 
NLPatchTST (mp) & 74.66 & 62.37 & 45.04 & 95.6 & 98.13 & 99.34 & 65.18 & 91.45 & 73.2 & 72.47 & 68.7 & 37.82 \\ 
SimMTM (s) & 78.66 & 66.18 & 50.72 & \doubleunderline{96.69} & 55.36 & 87.79 & \doubleunderline{73.08} & \doubleunderline{93.18} & \textbf{80.27} &  &  &  \\ 
SimMTM (p) & 85.32 & 67.73 & 52.95 & 96.41 & 55.38 & 87.98 & \textbf{74.2} & \textbf{93.4} & \doubleunderline{79.76} &  &  &  \\ 
Ci (s) & \textbf{88.7} & \textbf{69.0} & \textbf{55.79} & 96.11 & \doubleunderline{99.13} & \textbf{99.77} & 61.47 & 91.34 & 70.7 & 72.5 & \textbf{70.19} & \doubleunderline{40.36} \\ 
Ci (p) & \doubleunderline{87.46} & \doubleunderline{68.83} & \doubleunderline{55.5} & \textbf{96.75} & \textbf{99.14} & \underline{99.74} & 61.7 & 90.43 & 69.77 & \textbf{75.81} & \doubleunderline{70.14} & \textbf{40.6} \\ 
CiTrus (s) & 85.03 & 68.44 & 54.79 & 96.37 & \underline{99.12} & \doubleunderline{99.75} & 63.04 & 90.89 & 71.21 & 74.84 & 69.49 & 38.65 \\ 
CiTrus (p) & 83.86 & 67.18 & 52.71 & 96.45 & 98.82 & 99.58 & 63.26 & 91.37 & 71.09 & \underline{75.38} & \underline{69.77} & \underline{39.89} \\ 
CiTrus (fp) & \underline{87.42} & \underline{68.56} & \underline{55.05} & 96.48 & 98.95 & 99.69 & \underline{67.9} & \underline{92.44} & \underline{76.12} & 64.7 & 66.43 & 34.68 \\ 
CiTrus (mp) & 83.52 & 67.28 & 52.95 & \underline{96.51} & 98.57 & 99.37 & 62.95 & 90.23 & 70.64 & 72.82 & 68.56 & 38.07 \\ 
\hline
\end{tabular}
\caption{A comparison of the different model architectures for the high-data regime. We use the same evaluation method for each model, for EMG and FD-B we interpolate the data for fine-tuning, and for the other datasets we use our sliding window approach. The letters in brackets refer to how the model is trained; (s) is trained from scratch, (p) means it is pre-trained, (mp) means it uses multi-modal pre-training, and (fp) means it uses frequency pre-training. ACC, ROC, and PRC refer to the accuracy, area under the receiver operating characteristic, and the area under the precision recall curve, respectively. The best result for each metric is shown in bold, the second best result is double-underlined, and the third best result is single-underlined.}
\label{tab:high-data}
\end{table*}
\FloatBarrier

\clearpage
\end{document}